\documentclass[sigconf]{acmart}

\usepackage{booktabs} 

\usepackage[ruled]{algorithm2e} 

\SetAlFnt{\small}
\SetAlCapFnt{\small}
\SetAlCapNameFnt{\small}
\SetAlCapHSkip{0pt}

\usepackage{multirow}
\usepackage{caption}
\usepackage{subcaption}
\def\method{CondMDI}

\newcommand\blfootnote[1]{%
  \begingroup
  \renewcommand\thefootnote{}\footnote{#1}%
  \addtocounter{footnote}{-1}%
  \endgroup
}

\SetKwComment{Comment}{\quad $\triangleright$ }{ }
\copyrightyear{2024}
\acmYear{2024}
\setcopyright{acmlicensed}\acmConference[SIGGRAPH Conference Papers '24]{Special Interest Group on Computer Graphics and Interactive Techniques Conference Conference Papers '24}{July 27-August 1, 2024}{Denver, CO, USA}
\acmBooktitle{Special Interest Group on Computer Graphics and Interactive Techniques Conference Conference Papers '24 (SIGGRAPH Conference Papers '24), July 27-August 1, 2024, Denver, CO, USA}
\acmDOI{10.1145/3641519.3657414}
\acmISBN{979-8-4007-0525-0/24/07}

\usepackage{amsmath,amsfonts,bm}

\def\eqref#1{equation~\ref{#1}}

\def\1{\bm{1}}

\def\rvepsilon{{\bm{\epsilon}}}

\def\rvc{{\mathbf{c}}}

\def\rvm{{\mathbf{m}}}

\def\rvp{{\mathbf{p}}}

\def\rvr{{\mathbf{r}}}

\def\rvv{{\mathbf{v}}}

\def\rvx{{\mathbf{x}}}

\def\rvz{{\mathbf{z}}}

\def\vzero{{\bm{0}}}

\def\vmu{{\bm{\mu}}}

\def\mI{{\bm{I}}}

\def\mSigma{{\bm{\Sigma}}}

\DeclareMathAlphabet{\mathsfit}{\encodingdefault}{\sfdefault}{m}{sl}
\SetMathAlphabet{\mathsfit}{bold}{\encodingdefault}{\sfdefault}{bx}{n}

\def\gN{{\mathcal{N}}}

\newcommand{\meanp}[2]{\mathbb{E}_{#1} \left\lbrack #2 \right\rbrack}

\newcommand{\norm}[1]{\left\lVert#1\right\rVert}

\begin{document}

\title{Flexible Motion In-betweening with Diffusion Models}



\author{Setareh Cohan}
\affiliation{
    \institution{University of British Columbia}
    \country{Canada}
    }
\email{setarehc@cs.ubc.ca}

\author{Guy Tevet}
\affiliation{
  \institution{Tel-Aviv University}
  \country{Israel}
}
\affiliation{
    \institution{University of British Columbia}
    \country{Canada}
    }
\email{guytevet@mail.tau.ac.il}

\author{Daniele Reda}
\affiliation{
  \institution{University of British Columbia}
  \country{Canada}
}
\email{dreda@cs.ubc.ca}

\author{Xue Bin Peng}
\affiliation{
    \institution{Simon Fraser University}
    \country{Canada}
}
\affiliation{
    \institution{NVIDIA}
    \country{Canada}
}
\email{xbpeng@sfu.ca}

\author{Michiel van de Panne}
\affiliation{
    \institution{University of British Columbia}
    \country{Canada}
}
\email{van@cs.ubc.ca}
\begin{abstract}
Motion in-betweening, a fundamental task in character animation, consists of generating motion sequences that plausibly interpolate user-provided keyframe constraints. It has long been recognized as a labor-intensive and challenging process. We investigate the potential of diffusion models in generating diverse human motions guided by keyframes.
Unlike previous inbetweening methods, we propose a simple unified model capable of generating precise and diverse motions that conform to a flexible range of user-specified spatial constraints, as well as text conditioning. To this end, we propose Conditional Motion Diffusion In-betweening (\method{}) which allows for arbitrary dense-or-sparse keyframe placement and partial keyframe constraints while generating high-quality motions that are diverse and coherent with the given keyframes.
We evaluate the performance of \method{} on the text-conditioned HumanML3D dataset and demonstrate the versatility and efficacy of diffusion models for keyframe in-betweening.
We further explore the use of guidance and imputation-based approaches for inference-time keyframing and compare \method{} against these methods. 
\end{abstract}

%
%

\ccsdesc[500]{Computing methodologies~Machine learning}
\ccsdesc[500]{Computing methodologies~Animation}
%
%

\keywords{motion generation, motion in-betweening, diffusion models}

\begin{teaserfigure}
  \vspace{-.1cm}
  \centering
  \resizebox{.9\textwidth}{!}{
    \includegraphics[height=4cm,trim={0 30 0 0},clip]{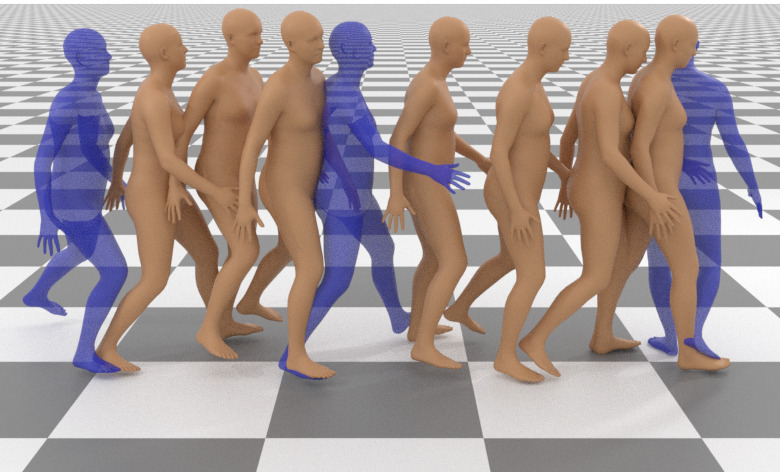}
  \includegraphics[height=4cm,trim={10 40 0 0},clip]{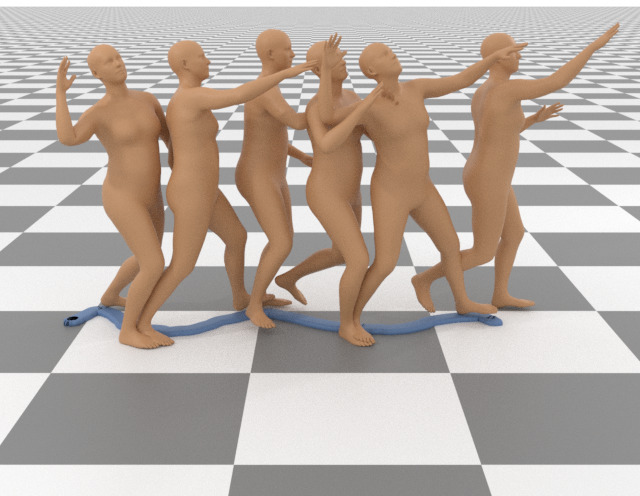}
      \includegraphics[height=4cm,trim={5 10 0 0},clip]{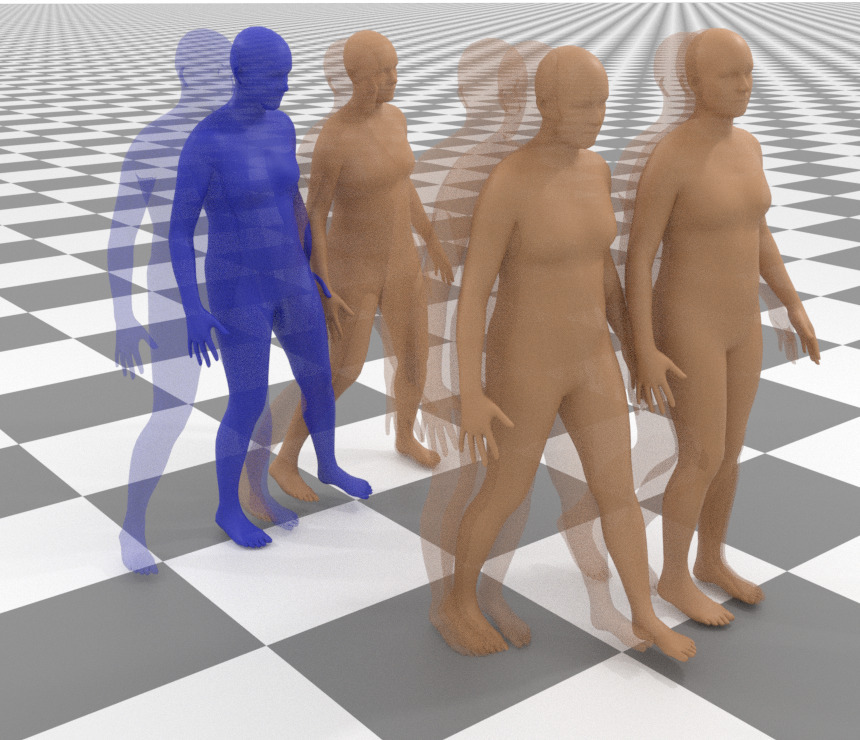}
    }
  \caption{Flexible motion in-betweening given a text prompt and spatio-temporally sparse keyframes. From left to right: 
  a) motion conditioned on sparse keyframes;
  b) motion conditioned on root trajectory and a "throwing" prompt;
  c) diverse motions generated for the same keyframes.}
  \label{fig:teaser}
\end{teaserfigure}
\maketitle

\section{Introduction}
\label{sec:intro}
Motion synthesis stands as a central challenge in computer animation, where the precise crafting of realistic movements is essential for conveying natural and lifelike behaviors. Keyframe in-betweening is a critical component of this process, but it is well known to be a demanding and time-consuming manual task.
\blfootnote{
\noindent Code and visualizations are available at \textbf{\href{https://setarehc.github.io/CondMDI/}
    {\texttt{setarehc.github.io/CondMDI/}}}.
}
Deep learning-based approaches have recently made significant progress on motion in-betweening, leveraging the availability of large-scale and high-quality motion capture datasets. Recurrent neural networks (RNNs) have been studied for the task of keyframe completion \cite{holden2016deep, zhang2018data, harvey2020robust, wang2022velocity}, however these RNN models can struggle to accurately model long-term dependencies.
Generative modeling techniques have also recently been applied to the task of motion in-betweening \cite{zhou2020generative, li2021task, he2022nemf}, with
transformer-based architectures modeling the long-term dependencies for keyframe motion completion~\cite{duan2021single, oreshkin2023motion}.

Most recently, diffusion-based models have demonstrated promising capabilities for generating diverse and realistic human motions \cite{tevet2023human, dabral2023mofusion, zhang2022motiondiffuse}. Diffusion models stand out for their ability to seamlessly incorporate constraints into the generation process, enabling precise control over the generated outputs. Notable examples include text-to-image generation using guidance \cite{nichol2021glide}, image completion using inpainting~\cite{lugmayr2022repaint, saharia2022palette}, and offline reinforcement learning with reward guidance \cite{janner2022planning}.

While diffusion models excel as robust conditional generation models, offering unique capabilities for inference-time conditioning, integrating spatial constraints, such as keyframes, into the motion generation process still has no standard solution. In this work, we present a unified and flexible method for motion in-betweening based on a masked conditional diffusion model called Conditional Motion Diffusion In-betweening (\method{}). This method trains on randomly sampled keyframes with randomly sampled joints, together with a mask that indicates the observed keyframes and features. This then offers significant flexibility in terms of number of keyframes and their placement in time, as well as partial keyframes, i.e., providing information for a subset of the joints.

Our key contribution is a simple and unified diffusion model for motion in-betweening, offering flexible inference-time conditioning. This model is trained by sampling from the space of all possible motion in-betweening scenarios. Our model accommodates temporally-sparse keyframes and partial pose specifications, alongside text prompts. This enables generation of high-quality motion sequences aligned with the specified constraints, while maintaining fast inference speed compared to alternative diffusion-based methods. We additionally provide experimental insights into alternative design choices, including imputation and reconstruction guidance methods.
\section{Related Work}
\label{sec:rw}
Kinematic methods for character animation have a long history. In the following, we first review longstanding data-driven methods, followed by more recent deep-learning based methods, and finally methods focusing specifically on motion in-betweening.

Since the advent of motion capture, numerous methods animate human movement by
temporally stitching together captured motion clips to meet user requirements.
Motion graphs can precompute feasible motion transitions~\cite{arikan2002interactive, lee2002interactive, kovar2002motion}, which can then be used to synthesize motions via search~\cite{lee2002interactive, kovar2002motion}, dynamic programming~\cite{arikan2003motion, hsu2004example, pullen2002motion}, path planning~\cite{safonova2007construction}, and reinforcement learning~\cite{lee2004precomputing, mccann2007responsive, lo2008real}. Motion matching~\cite{buttner2015motion} is a related method that searches for animation frames that best fit the current context.
Motion blending methods further allow for interpolation of motions. Radial basis function (RBF) kernels have been used to interpolate motions of the same class~\cite{rose1998verbs, rose2001artist}. 
Some work cluster similar motions~\cite{kovar2004automated,beaudoin2008motion} while others develop statistical models that allow the original data to be discarded, e.g., ~\cite{mukai2005geostatistical,chai2007constraint}.

Deep learning methods have proliferated through animation. Human motion synthesis models are typically trained using large collections of motion capture data~\cite{AMASS:ICCV:2019,mixamo,guo2022generating}. 
A large class of parametric models have been proposed for motion modeling, such as RNNs~\cite{fragkiadaki2015recurrent, ghosh2017learning, li2017auto, aksan2019structured},
autoencoders~\cite{holden2015learning, holden2016deep,guo2020action2motion, ling2020character,li2021task,zhang2023t2m}, and 
GANs~\cite{ahn2018text2action, ghosh2021synthesis}. Inspired by the success of Flow-based models for image synthesis~\cite{dinh2014nice}, auto-regressive normalizing networks for motion sequence modeling have also been proposed~\cite{henter2020moglow}.

More recently, denoising diffusion models have been widely utilized for motion synthesis~\citep{tevet2023human,zhang2022motiondiffuse,dabral2023mofusion,kim2022flame}. Diffusion-based methods have proved to have a high capacity for modeling the complex distributions associated with motion data and have enabled new types of control over the motion generation. Notable instances are trajectory and joint control by PriorMDM~\cite{shafir2023human}, GMD~\cite{karunratanakul2023guided} and OmniControl~\cite{xie2023omnicontrol}; Multi-person interactions by ComMDM~\cite{shafir2023human}, and InterGen~\cite{liang2023intergen}. The flexibility of diffusion models was also demonstrated for non-human motion synthesis in MAS~\cite{kapon2023mas} and SinMDM~\cite{raab2023single}. 

Motion in-betweening generates a full motion sequence given a set of keyframes with their associated timing. Motion in-betweening can be cast as a motion planning problem, capable of synthesizing fairly complex motions~\cite{arikan2002interactive, beaudoin2008motion, levine2012continuous, safonova2007construction}. Effective data structures such as motion graphs made search and optimization more efficient~\cite{kovar2002motion, min2012motion, shen2017posture}. These methods suffer from memory and scalability issues as they either require maintaining a motion database in memory or performing search and optimization at run-time~\cite{harvey2020robust}. Deep learning can overcome these limitations by utilizing large datasets for training while having a fixed computation budget at run-time~\cite{harvey2020robust}. Due to the temporal nature of the task, RNN-based methods have dominated the field \cite{zhang2018data, harvey2018recurrent, harvey2020robust}. RNN-based models can struggle with long-term dependencies and are thus often limited to generating shorter transition animations. Unlike auto-regressive models, Transformer-based~\cite{vaswani2017attention} models predict the entire motion trajectory at once~\cite{duan2021single, oreshkin2023motion, qin2022motion}. VAEs and GANs have also been applied to motion in-betweening~\cite{zhou2020generative, li2021task, he2022nemf}. A key limitation of these methods is the models are generally limited to fixed keyframe patterns. 

Diffusion-based methods allow for keyframe-based control, e.g., via imputation and inpainting methods.
However when methods such as MDM~\cite{tevet2023human} are presented with inpainted
full joint trajectories, the motions exhibit very significant foot sliding and
unnatural movements to satisfy the constraints. PriorMDM~\cite{shafir2023human} suggests fine-tuning MDM with the observed trajectory of interest. Both methods do not allow for global or sparse-in-time constraints due to a relative-to-previous-frame representation for global root-joint translation and orientation. 
GMD~\cite{karunratanakul2023guided} supports sparse-in-time keyframes, but only allows for specification of the pelvis position alone rather than the full pose. Hence, sparse keyframes in this work refer to sparse positions of the root joint and their method solves a goal-reaching task rather than keyframe in-betweening. 
GMD proposes a two-stage pipeline: root trajectory synthesis, then full-body motion generation conditioned on the synthesized root trajectory. It relies on inference-time imputation and guidance and a specialized emphasis-projection technique to increase the importance of observed keyframes.

Closest to our own work, OmniControl~\cite{xie2023omnicontrol} introduces controllable motion generation with a full-pose spatial conditioning signal, representing global positions of joints over time. While intended for joint control rather than keyframe in-betweening, it supports multiple-joint keyframes and allows for full keyframe conditioning via 3D joint positions, but not joint rotations. OmniControl uses MDM as its diffusion backbone and utilizes a trainable copy of the Transformer encoder of MDM to embed the keyframe signal and later adds them to the attention layers of MDM. In addition to requiring this separate embedding module, OmniControl relies on repeated guidance application to further enforce the constraints. In addition to adding complexity. These features notably increase the inference time of OmniControl compared to other diffusion-based motion generation models.
\section{Background}
\label{sec:background}
In this section, we first review diffusion probabilistic models for motion generation which we refer to as motion diffusion models. Next, we provide an overview of different conditioning approaches applicable to diffusion models for conditional motion generation.

\begin{figure*}[t!]
    \centering
        \includegraphics[width=0.875\textwidth]{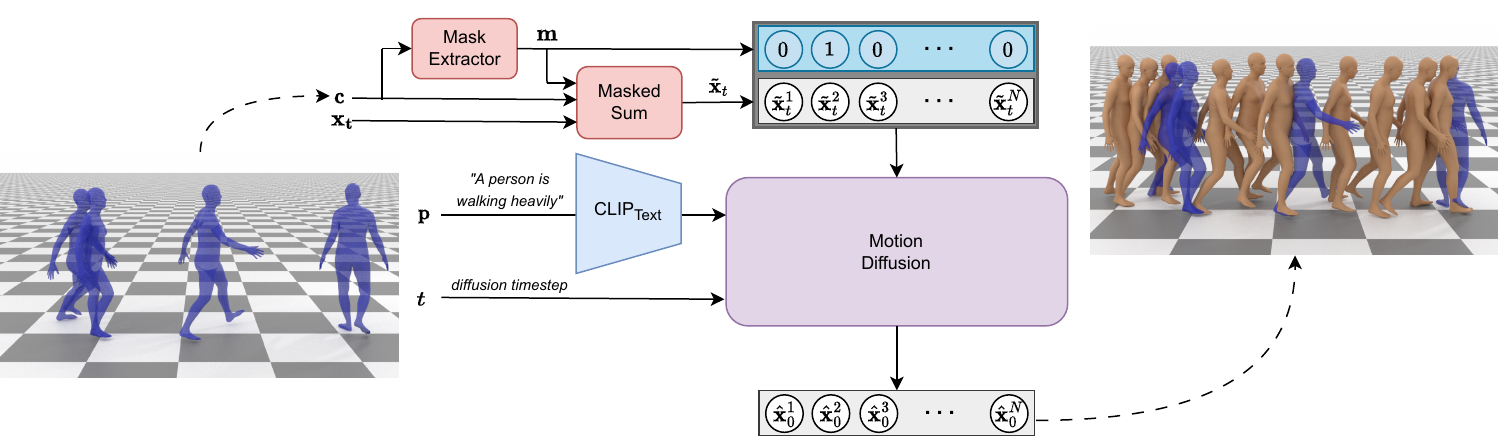}
        \caption{{\bf Conditional Motion Diffusion In-betweening (CondMDI) overview}. The model is fed a noisy motion sequence $\rvx_t$, the diffusion step $t$, a text prompt $\rvp$, and a keyframe control signal $\rvc$. Text prompt $\rvp$ is first fed into a CLIP-based \cite{radford2021learning} textual embedder before being fed into the motion diffusion model which is based on GMD \cite{karunratanakul2023guided}. Mask Extractor module extracts the binary mask and the Masked Sum module performs the masked addition $\tilde{\rvx}_t = \rvm \odot \rvc + (\mathbf{1} - \rvm) \odot \rvx_t$ and the gray box around $\tilde{\rvx}_t$ and $\rvm$ indicates concatenation of the two.}
        \label{fig:overview}
\end{figure*}

\subsection{Human Motion Generation with Diffusion Models}
\label{subsec:mdm}
Diffusion models have shown incredible capabilities as generative models \cite{sohl2015deep, ho2020denoising, song2019generative} and they are the backbone of current state-of-the-art (SOTA) image synthesis models, such as Imagen \cite{saharia2022photorealistic} and DALL-E2 \cite{ramesh2022hierarchical}. Viewing motion synthesis as a sequence generation problem, diffusion probabilistic models have been recently applied to generate the entire motion sequence at one go \cite{tevet2023human}.

Given a motion dataset, diffusion models add small amounts of Gaussian noise to the samples $\rvx_0 \sim q(\rvx_0)$ in $T$ steps such that the marginal distribution at diffusion step $T$ is $q(\rvx_T) \approx \mathcal{N}(\rvx_T; \vzero, \mI)$. This is known as the \emph{forward process} and is formulated as:
\begin{align}
    &q(\rvx_t | \rvx_{t-1}) := \gN(\rvx_t; \sqrt{1-\beta_t}\rvx_{t-1},\beta_t\mathbf{I})
\end{align}\label{eq:forward_process}
where $t$ is the diffusion step and $\beta_{1,...,T}$ is a fixed variance schedule indicating the amount of noise. To generate samples conditioned on text prompts $\rvp$, diffusion models learn the \emph{reverse process} of removing noise from $\rvx_t$ starting from pure Gaussian noise $\rvx_T$:
\begin{align}
    p_\theta(\rvx_{t-1} | \rvx_t, \rvp) := \mathcal{N}(\rvx_{t-1}; \vmu_\theta(\rvx_t,t,\rvp), \mSigma_t)
    \label{eq:reverse_process}
\end{align}
where $\theta$ are the model parameters, and $\mSigma_t$ is the untrained time-dependable covariance set according to the variance schedule.

Most motion diffusion models use the sample-estimation reparameterization and directly predict the clean sample estimate $\hat{\rvx}_0$ instead of the mean estimate $\vmu$. In this case the final objective to optimize the diffusion model $G_\theta(\rvx_t, t, \rvp)$ is:
\begin{align}
    \mathcal{L} := \meanp{(\rvx_0,\rvp) \sim q(\rvx_0, \rvp), t \sim [1, T]}{
        \norm{
            \rvx_0 - G_\theta(\rvx_t, t, \rvp)
        }^2
    }.
    \label{eq:loss}
\end{align}
Given the sample estimate $\hat{\rvx}_0$, the mean estimate $\tilde{\vmu}$ is computed as:
\begin{align}
    \tilde{\vmu}_\theta(\rvx_t,t,\rvp,\hat{\rvx}_0) := \frac{\sqrt{\bar{\alpha}_{t-1}} \beta_t}{1 - \bar{\alpha_t}} \hat{\rvx}_0 + \frac{\sqrt{\alpha_t} (1 - \bar{\alpha}_{t-1})}{1 - \bar{\alpha}_t} \rvx_t
    \label{eq:tilde_mu}
\end{align}
with $\alpha_t:=1-\beta_t$ and $\bar{\alpha_t} := \prod_{s=1}^{t} \alpha_t$.

To allow for some flexibility over the relative strength of the condition, \emph{classifier-free guidance} is typically used with text-conditioned motion generation. Classifier-free guidance proposes to train an unconditional model $G_\theta(\rvx_t, t)$ jointly with $G_\theta(\rvx_t, t, \rvp)$ by setting $\rvp=\emptyset$ for a fraction of training samples, e.g., $10\%$.
The weighted combination of the two predictions is output at inference time:
\begin{align}
    G_\theta(\rvx_t, t, \rvp) = {G_\theta(\rvx_t, t, \emptyset)} + w \left( {G_\theta(\rvx_t, t, \rvp)} - {G_\theta(\rvx_t, t, \emptyset)} \right)
\end{align}\label{eq:classifier-free-guidance}
where $w$ helps trade-off between fidelity to the text prompt and diversity among samples.

\subsection{Conditional Motion Generation with Diffusion Models}
\label{subsec:cond_motion}
Incorporating spatial constraints such as keyframes into motion diffusion models can be done through two distinct approaches: 1) Training a diffusion model explicitly trained given the spatial conditioning signal as input, and 2) Leveraging a pre-trained motion diffusion model with inference-time imputation and guidance.

\paragraph{Explicit Conditional Models}
In this approach, the spatial conditioning signal $\rvc$ will be used as an additional input to the motion diffusion model $p_\theta(\rvx_{t-1}|\rvx_t, \rvp, \rvc)$. The model is trained to learn this conditional distribution. 

\paragraph{Inference-time Imputation}
Diffusion models allow for manipulating the generated samples to satisfy certain conditions at inference time. If the spatial conditioning signal $\rvc$ is an observed part of the desired motion sample (e.g. partial keyframes), \emph{imputation} or \emph{inpainting}~\cite{lugmayr2022repaint} can be used to generate samples that adhere to this observation. This is done by replacing the output of the pre-trained diffusion model $\rvx_t$ with the noisy version of the observation $\rvc$ over the observation mask $\rvm$ at each diffusion step $t$.

\paragraph{Inference-time Guidance}
In addition, \emph{guidance} can also be used to push the samples towards the desired spatial condition. Let $\mathcal{J}(\rvx)$ be a loss function defining how much motion sequence $\rvx$ deviates from $\rvc$. With guidance, gradients of $\mathcal{J}$ can be used to guide the output of the diffusion model towards minimizing this loss~\cite{dhariwal2021diffusion, janner2022planning}. \emph{Reconstruction guidance}~\cite{ho2022video} is a special form of guidance that operates on sample estimates $\hat{\rvx}_0$ and is used to improve the cohesion between observations and the generated motion. To do so, the loss function $\mathcal{J}$ is defined as the MSE over the observed spatial constraints and the diffusion model's predictions for the observations. At every denoising step $t$, model's predictions for the unobserved parts will be adjusted as below:
\begin{align}
     \hat{\rvx}_{0,t}^{\texttt{p}} = \hat{\rvx}_{0,t}^{\texttt{p}} - \frac{w_r \sqrt{\bar{\alpha}_t}}{2} \nabla_{\rvx^p_t}\left\| {c - \hat{\rvx}_{0,t}^{\texttt{o}}}\right\|^2
\end{align}
where superscripts $\texttt{o}$ and $\texttt{p}$ refer to the observed and predicted parts of the sequence respectively, and $w_r$ is guidance weight.
\section{Method}
\label{sec:method}
When performing motion keyframing, our goal is to produce realistic motions that adhere to a set of spatio-temporarily sparse input keyframes while maintaining coherence between these observed keyframes and the entirety of the generated motion sequence. In this section, we first provide the detailed problem setup. Then we provide a discussion of the motion data representation and how it affects our keyframe in-betweening method \method{}, followed by a detailed description of \method{}.

\subsection{Problem Definition}
\label{sec:problem_def}
Given a text prompt $\rvp$, observation control signal $\rvc \in \mathbb{R}^{N \times J \times D}$, our goal is to generate a human motion trajectory $\rvx = \{\rvx^i\}_{i=1}^N \in \mathbb{R}^{N \times J \times D}$ where $N$ is the number of frames. The pose of $i$-th frame $\rvx^i \in \mathbb{R}^{J \times D}$ is represented by a $D$-dimensional feature vector for the pose of $J$ joints. For our task of keyframe in-betweening, the control signal $\rvc$ contains only an observed subset of $ k \leq N$ keyframes (temporal sparsity) for a subset of $j \leq J$ joints (spatial sparsity).

\subsection{Motion Representation}
\label{subsec:motion_representation}
Common motion representations divide each motion sequence into two parts: \emph{local motion} containing the pose of the skeleton relative to the root at every frame, and \emph{global motion} containing the global translations and rotations of the root joint relative to the previous frame \cite{he2022nemf, karunratanakul2023guided}. Referring back to the problem definition above, a small portion out of the $D$ features includes the global orientation of the root with respect to the previous frame, and the rest of the features represent the local pose with respect to the root joint. Since the root joint positions are represented as relative positions with respect to the previous frame, incorporating temporarily sparse spatial constraints such as sparse keyframes, adds an additional challenge to the sparse keyframing problem. Thus, we address this challenge by converting the relative orientation of the root to global coordinates and use this global-root representation for our model. Detailed description of this conversion is available in Appendix~\ref{supp:root_represenation}.

\subsection{Conditional Motion Diffusion In-betweening}
\label{subsec:inbetweening}
We model the conditional reverse posterior $p_\theta(\rvx_{t-1} | \rvx_t, \rvp, \rvc)$ with an explicit conditional diffusion model which takes the keyframe conditioning signal $\rvc$ as input alongside the noisy motion sample $\rvx_t$ and the text prompt $\rvp$. An overview of our approach is represented in Figure~\ref{fig:overview}. To incorporate the keyframe information, following~\cite{harvey2022flexible}, we adopt a straightforward approach and replace the noisy sample $\rvx_t$ with the observed partial keyframes $\rvc$ at every observed frame and joint. To provide the model with an indication of which features are observed, we concatenate the resulting masked sample $\rvx_t$ with the observation mask as input to the diffusion model. The observation mask $\rvm \in \mathbb{R}^{N \times J \times D}$ is a binary mask with ones over the observed frames and joints and zero everywhere else, defined based on the keyframe signal $\rvc$. To allow for flexible keyframe conditioning at inference-time, our model is trained with randomly sampled partial keyframes. Algorithm~\ref{alg:training_cond} shows an overview of the training procedure of our conditional method. $\texttt{Random Mask Generator}$ is the procedure in which the number of keyframes $k$ is first sampled within the length of the motion sequence, and then these $k$ keyframes are randomly picked out of all the frames in the sequence. To provide additional flexibility over the joints, this method is extended to additionally sample the number of observed joints $j$, and then randomly pick the observed joints out of all $J$ joints. Note that we set keyframe conditioning signal $\rvc$ to $\emptyset$ for $10\%$ of training samples to make \method{} better suited for unconditioned motion generation at inference time. Our proposed conditioning method can be applied to any backbone text-conditioned motion diffusion model $G_\theta$, and we choose to use the motion diffusion model of GMD~\cite{karunratanakul2023gmd} as our backbone diffusion model. For more details about the network architecture, refer to Appendix~\ref{supp:architecture}.
$\odot$ is the element-wise product and $\left < \right >$ are used to denote concatenation. The sampling procedure of our conditional method is available in Algorithm~\ref{alg:sampling_cond}.

\begin{algorithm}[t]
  \caption{Training} \label{alg:training_cond}
  \small
    \Repeat{converged}{
      $(\rvx_0, \rvp) \sim q(\rvx_0, \rvp)$ \\
      $\rvm \sim \texttt{Random Mask Generator}$ \\
      $\rvp \leftarrow \emptyset$ with probability $10\%$ \Comment{Classifier-free Guidance}
      $\rvc \leftarrow \emptyset$ with probability $10\%$ \Comment{Unconditioned Generation}
      $t \sim \mathrm{Uniform}(\{1, \dotsc, T\})$ \\
      $\rvepsilon\sim\mathcal{N}(\vzero,\mI)$ \\
      $\rvx_t = \sqrt{\bar{\alpha_t}} \rvx_0 + \rvepsilon \sqrt{1-\bar{\alpha_t}}$ \\
      $\rvx_t = \rvm \odot \rvx_0 + (1 - \rvm) \odot \rvx_t$ \\
      $\rvx_t = \left< \rvx_t, \rvm \right>$ \\
      Take gradient descent step on \\
      $\qquad \nabla_\theta \left\| \rvx_0- G_\theta(\rvx_t, t, \rvp) \right\|^2$
      }
\end{algorithm}

\hfill

\begin{algorithm}[t]
  \caption{Sampling} \label{alg:sampling_cond}
  \small
    \textbf{Require:} Guidance scale $w$  \\
    \textbf{Require:} Text prompt $\rvp$  \\
    \textbf{Require:} Keyframe signal $\rvc$ and observation mask $\rvm$ \\
    $\rvx_T \sim \mathcal{N}(\vzero, \mI)$ \\
    \For{$t=T, \dotsc, 1$}{
      $\rvz \sim \mathcal{N}(\vzero, \mI)$ if $t > 1$, else $\rvz = \vzero$ \\
      $\rvx_t = \rvm \odot \rvc + (1 - \rvm) \odot \rvx_t$ \\
      $\rvx_t = \left< \rvx_t, \rvm \right>$ \\
      $\hat{\rvx}_0 = G_\theta(\rvx_t,t,\emptyset) + w \left ( G_\theta(\rvx_t, t, \rvp) - G_\theta(\rvx_t,t,\emptyset) \right)$ \\
      $\hat{\vmu} = \tilde{\vmu}(\hat{\rvx}_0, \rvx_t)$ \\
      $\rvx_{t-1} = \hat{\vmu} + \sigma_t \rvz$ \\
    }
    \textbf{return} $\rvx_0$
\end{algorithm}

\section{Implementation and Evaluation Metrics}
\label{sec:experimental_setup}

Our method is evaluated on the human motion generation task conditioned on text prompts and a variety of keyframe control signals. In particular, we evaluate the performance of our method on text-conditioned motion generation given sparse keyframes. We also compare against inference-time conditioning methods for the task of in-betweening, including both imputation and imputation combined with reconstruction guidance. Finally, we evaluate our model on a wide range of conditioning signals to demonstrate the capabilities of our model beyond simple keyframes.

\subsection{Dataset}
Our model is evaluated on the HumanML3D~\cite{guo2022generating} dataset which contains 14,646 text-annotated human motion sequences taken from the AMASS~\cite{mahmood2019amass} and HumanAct12~\cite{guo2020action2motion} datasets. Motion sequences from this dataset have variable lengths where the average motion length is 7.1 seconds and motions are padded with zeros to be a fixed length of 196 frames with a framerate of $20$ fps. In this dataset, motion at every frame is represented by a $263$-dimensional feature vector consisting of the relative root joint translations and rotations, plus the local pose including the joint rotations and joint positions with respect to the root joint. Detailed description of the data representation is available in Appendix~\ref{supp:represenation}.\\

Sparse keyframes need to be defined with global translation and orientation of the root joint. To make conditioning of diffusion models on such global keyframes more straight-forward, we first convert the dataset to have global orientations for the root joint. For each frame, this is simply done by cumulatively summing the translation and rotation of the root joint up to its previous frame. \method{} assumes similar dimensionality for the keyframe signal and motion signal. Consequently, for each observed frame and joint, \method{} requires all corresponding features out of 263. Additionally, as motions in the dataset are represented as root motion and pose with respect to the root, partial keyframes always include the root joint. Further details on this can be found in Appendix~\ref{supp:keyframe_signal_details}.

\subsection{Evaluation Metrics}
For the task of conditional motion generation, we adopt the evaluation protocol from~\citet{guo2022generating}.
They suggest a set of neural metrics calculated in a mutual text-motion latent space based on pre-trained encoders.
This includes \emph{Fréchet Inception Distance (FID)} score, which measures the distance between the distribution of ground-truth and generated motions in the latent space of a pre-trained motion encoder. \emph{R-Precision} measures the proximity of the motion to the text it was conditioned on, and \emph{Diversity} measures the variability within the generated motion. The full description of these metrics is available in Appendix~\ref{supp:metrics}.
In addition, we adopt the \emph{Foot Skating Ratio} and the \emph{Keyframe Error} metrics from~\citet{karunratanakul2023guided}. The prior measures the proportion of frames in which either foot skids more than a certain distance (2.5 cm) while maintaining contact with the ground (foot height $<$ 5 cm). The latter measures the mean distance between the generated motion root locations and the keyframe root locations at the keyframe motion steps.

\subsection{Implementation Details}
For our baseline diffusion model, we adopt the motion diffusion model of GMD~\cite{karunratanakul2023guided}, which uses a UNet architecture with AdaGN~\cite{dhariwal2021diffusion}. Our model uses the sample-estimation parameterization of DDPMs~\cite{ho2020denoising} with $T=1000$ diffusion steps during training and inference. Similar to GMD, we use the pre-trained CLIP model to encode the text prompts~\cite{radford2021learning}. For more implementation details, refer to Appendix~\ref{supp:implementation_details}.
\begin{table} 
    \caption{\textbf{Text-to-motion evaluation} on the HumanML3D test set.}
    \footnotesize
    \centering
    \resizebox{0.99\columnwidth}{!}{
    \begin{tabular}{lcccc}
        \toprule
        & FID $\downarrow$~ & \multicolumn{1}{p{1.7cm}}{\centering R-precision $\uparrow$ \\ (Top-3)} & Diversity $\rightarrow$\\
        \midrule
        Real & 0.002 & 0.797 & 9.503 \\
        \midrule
        JL2P \cite{ahuja2019language2pose} & 11.02 & 0.486 & 7.676 & \\
        Text2Gesture \cite{bhattacharya2021text2gestures} & 7.664 & 0.345 & 6.409 & \\
        T2M \cite{guo2022generating} & 1.067 & 0.740 & 9.188 & \\
        MotionDiffuse \cite{zhang2022motiondiffuse} & 0.630 & \textbf{0.782} & \underline{9.410} & \\
        MDM & 0.556 & 0.608 & \textbf{9.446} & \\
        MLD \cite{chen2023executing} & 0.473  & \underline{0.772} & 9.724 & \\
        PhysDiff \cite{yuan2023physdiff} & 0.433  & 0.631 & - & \\
        {GMD $\rvx^\text{proj}$} & \textbf{0.235} & 0.652 & 9.726 \\
        \midrule
        {Ours} & \underline{0.2538} & 0.6450  & 9.7489\\
        \bottomrule
    \end{tabular}}
    \label{table:result_unconditional}
\end{table}
\begin{table}[t]
    \caption{\textbf{Quantitative results for different keyframes} on the HumanML3D test set. $K \in \{1, 5, 20\}$ means number of keyframes randomly placed along the motion trajectory. \textit{Root Joint} and \textit{VR Joints} mean conditioning on the root joint trajectory and the head and both wrist joints repectively.}
    \centering
    \resizebox{0.99\columnwidth}{!}{
    \begin{tabular}{c|ccccc}
        \toprule
        Conditioning & FID $\downarrow$ & \multicolumn{1}{p{1.9cm}}{\centering R-precision $\uparrow$ \\ (Top-3)} & Diversity $\rightarrow$ &   \multicolumn{1}{p{1.8cm}}{\centering Foot skating \\ ratio $\downarrow$ }
        & \multicolumn{1}{p{1.6cm}}{\centering Keyframe err $\downarrow$} \\
        \midrule
        Random K=1  & 0.1551 & 0.6787 & 9.5807 & 0.0936  & 0.3739  \\
        Random K=5  & 0.1731 &  0.6823 & 9.3053 & 0.0850  & 0.1789 \\
        Random K=20  & 0.2253 & 0.6821 & 9.1151 & 0.0806  & 0.0754  \\
        \midrule
        Root Joint & 0.2474  & 0.6752 & 9.4106 & 0.0854 & 0.0525 \\
        \midrule
        VR Joints & 0.2969 & 0.6842 & 9.0659 & 0.0794  & 0.0422  \\
        \bottomrule
        \end{tabular}}
\label{table:keyframing_schemes}
\end{table}

\section{Results}
\label{sec:results}
In this section, we present our empirical findings. In Sections~\ref{subsec:keyframes} and ~\ref{subsec:partial_keyframes}, we provide qualitative samples for sparse-in-time and sparse-in-time-and-joints keyframes. In Section~\ref{subsec:uncond_synth} we evaluate the performance of \method{} on the task of text-conditioned motion synthesis without any keyframe conditioning. Section~\ref{subsec:eval} contains evaluation results of \method{} on the text-and-keyframe conditioned motion generation task. Finally, Section~\ref{subsec:ablations} shows the ablation results. For additional results on sample diversity and text-conditioning, refer to Appendix~\ref{supp:additional_results}.
In the qualitative samples, generated and observed keyframes are shown in yellow and blue unless otherwise stated. Additionally, in all tables, \textbf{bold} indicates best result, \underline{underline} indicates second best, and $\rightarrow$ indicates that closer to real is better.

\subsection{Sparse Keyframe In-betweening}
\label{subsec:keyframes}
\begin{figure}[t]
\resizebox{0.8\linewidth}{!}{
\centering
\includegraphics[width=0.48\linewidth]{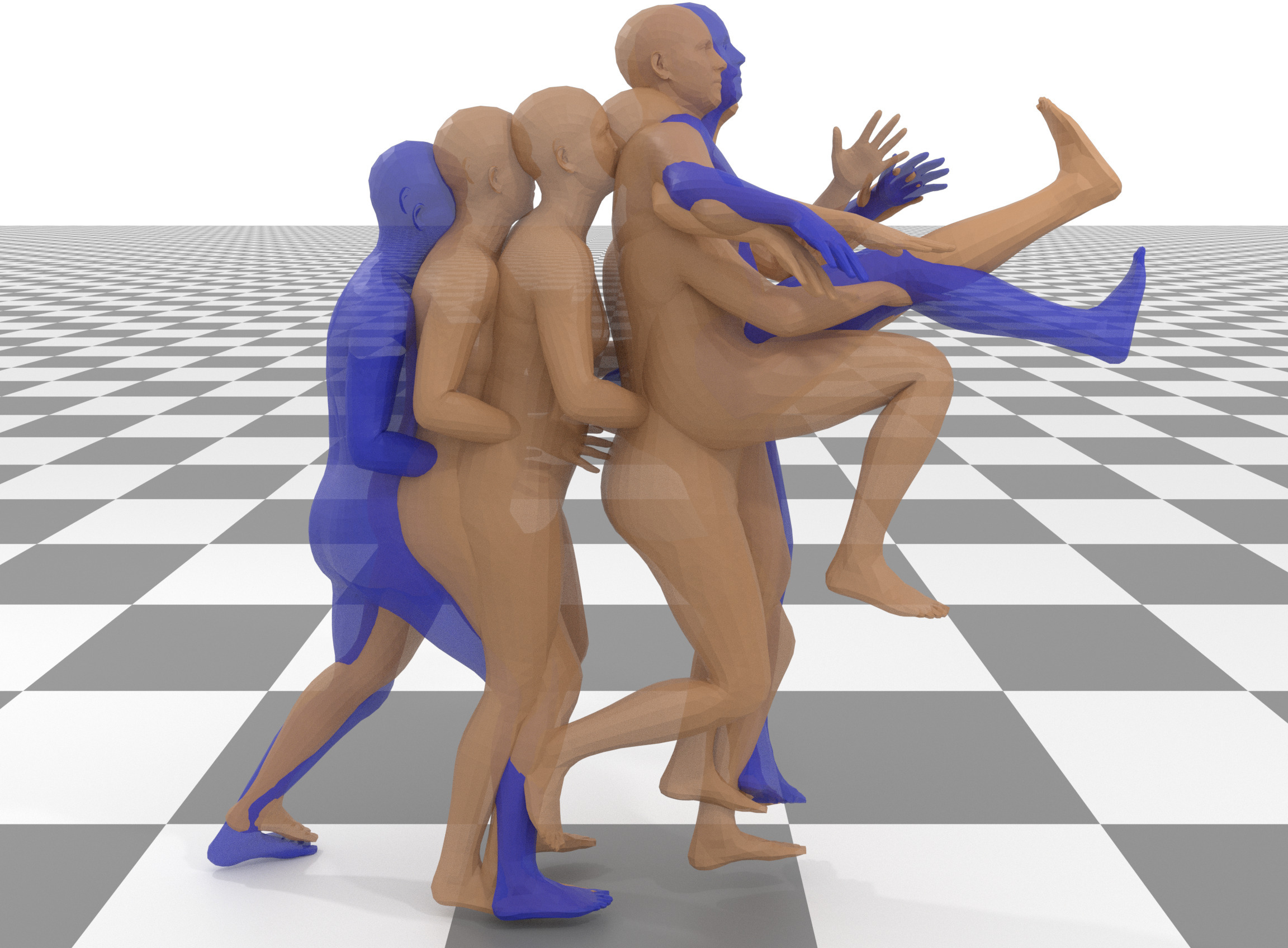}
\includegraphics[width=0.48\linewidth]{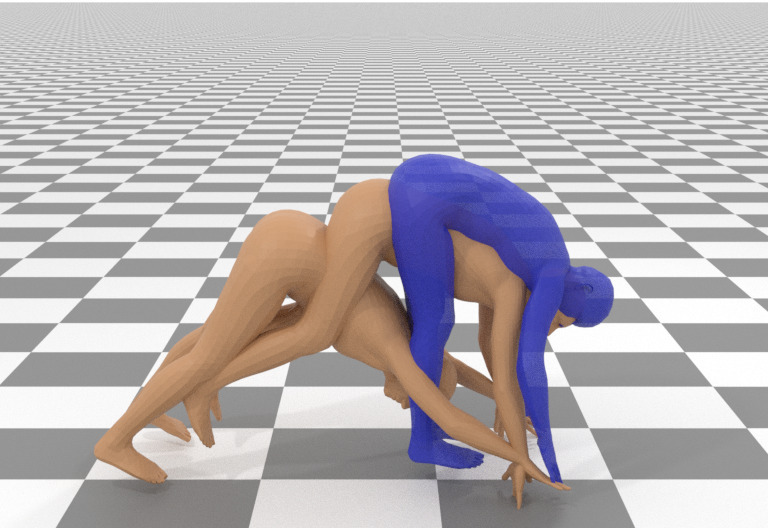}
}
    \caption{Our model is capable of generating high-quality motions in hard moves such as a karate kick or a yoga sun salutation pose. Check the video for the full motions.}
    \label{fig:karate-yoga}
\end{figure}

First, we evaluate the performance of our method on the task of sparse keyframe in-betweening, a primary focus of our model. For the classical case of sparse keyframe in-betweening, we first evaluate our model by creating samples using sparse keyframes provided at fixed transitions of $T$ frames. {Figure~\ref{fig:karate-yoga} shows that the model is capable of generating high quality motions from sparse keyframes placed every $T=20$ frames, even on dynamic and complex movements such as karate and yoga.} {Our qualitative results show that \method{} can generate smooth and high quality samples that
are consistent with the input keyframes, even with spacing of over 40 frames.} As a more general sparse keyframing approach, instead of specifying keyframes evenly spaced in time, we provide $K$ frames randomly spaced in time.

\subsection{Partial Keyframe In-betweening: Joint Control}
\label{subsec:partial_keyframes}
\begin{figure}[t]
\centering%
\includegraphics[width=0.55\linewidth,trim=20 150 30 100,clip]{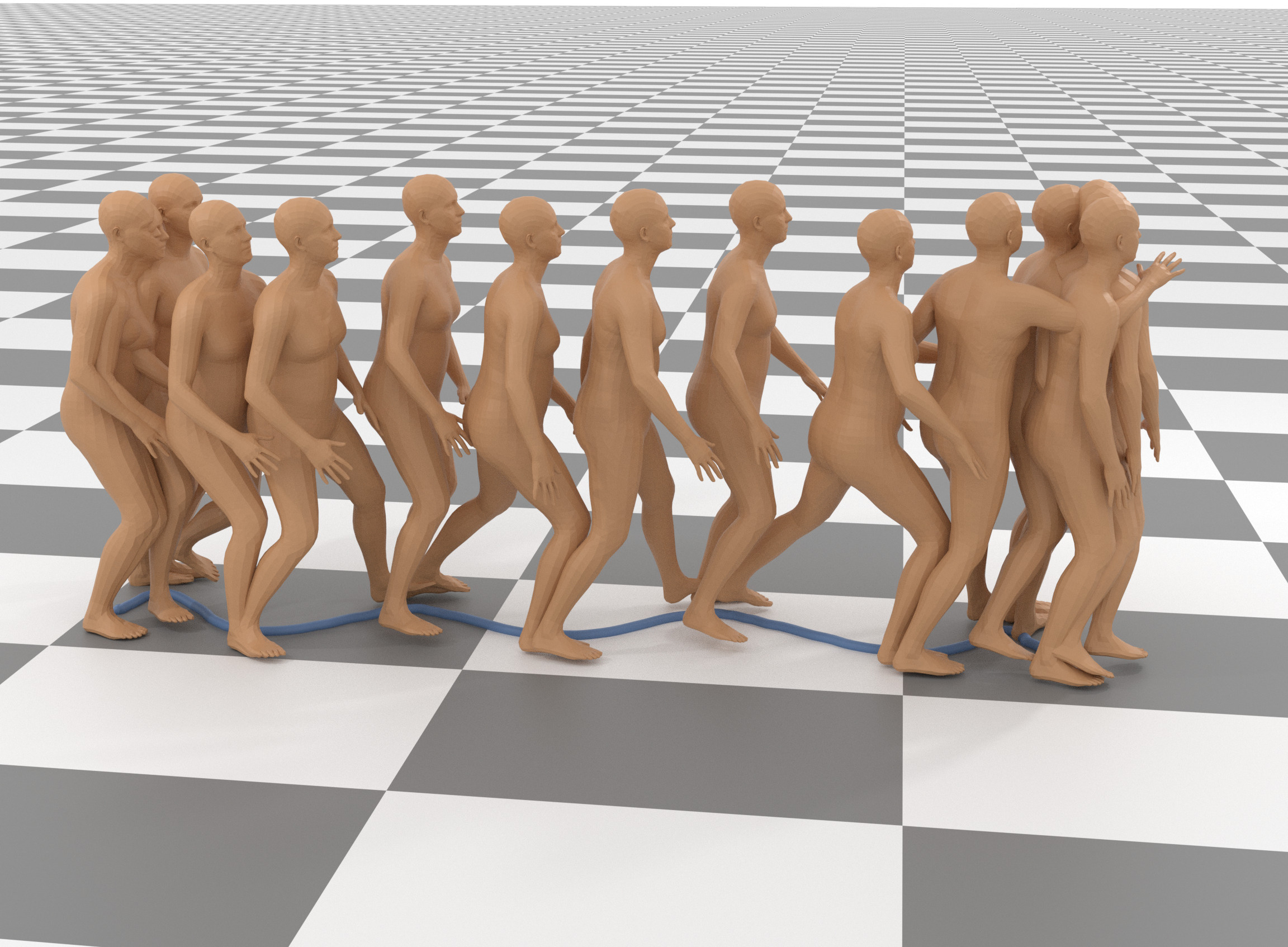}~%
\includegraphics[width=0.33\linewidth, trim=30 30 50 15,clip]{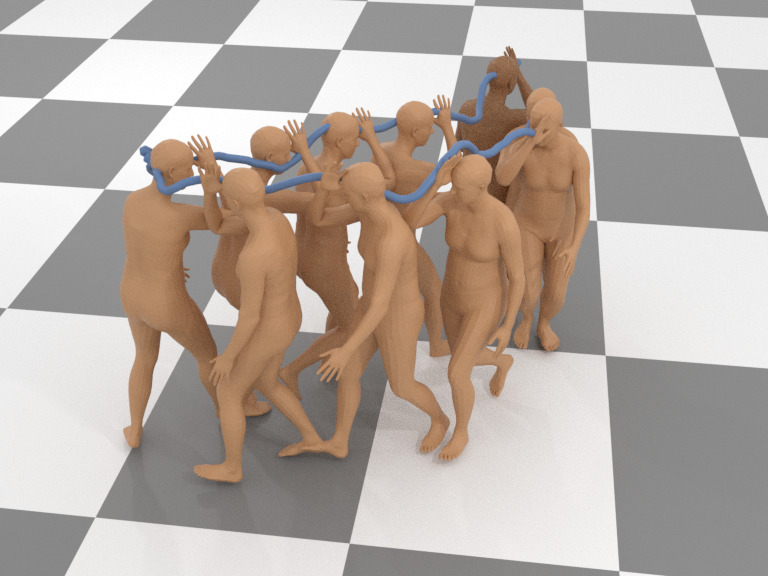}%
\caption{A walking motion conditioned only on the root joint (left) and only on the right wrist (right).}
\label{fig:root-joint-conditioning}
\end{figure}

To further test the capabilities of our model, we define spatially-sparse keyframes, i.e. keyframes that contain a subset of the joints. Our model demonstrates good performance even when provided with a single joint trajectory. Figure~\ref{fig:root-joint-conditioning} shows examples of the model provided with only the root joint trajectory (projected on the ground in the left figure), or with only the right wrist joint. The sample follows the input trajectory closely with natural and smooth motions. Partial keyframes also allow for other useful applications, such full-body motion reconstruction from sparse VR headsets, consisting of only the head, left wrist and right wrist joints. In the supplementary video, we show that our model is able to generate complex lower-body motions only from this sparse input.

\subsection{Unconditioned Synthesis}
\label{subsec:uncond_synth}
{In Table~\ref{table:result_unconditional}, we demonstrate the performance of \method{} on the task of text-conditioned motion synthesis. This table is added as a reference to interpret the values of the rest of the quantitative evaluations, in which \method{} also observes input keyframes. Conditioning the same model on keyframe
information, should ideally lead to superior performance,
as the space of solutions becomes more restricted. However, in practice, incorporating conditioning signals comes with unique challenges, generally leading to worse performance for conditioned models in terms of motion quality metrics. Therefore, a decrease in the average keyframe error while maintaining or improving the rest of the metrics shows the effectiveness of a model conditioned on keyframes.}

\subsection{Evaluation}
\label{subsec:eval}
For a grounded evaluation, we test our model by computing quantitative results for a range of keyframing schemes. Results are shown in Table~\ref{table:keyframing_schemes}. For the three cases {with $K \in \{1, 5, 20\}$} randomly placed full keyframes, we can see that as the number of observed keyframes increases, the average error of the keyframes decreases due to denser conditioning, providing a stronger influence on the model. However, increasing keyframes results in worse FID values, possibly because denser signals may constrain the model too much, leading to performance degradation. Overall, all these cases demonstrate performance comparable to or better than unconditional synthesis for the motion quality metrics, while exhibiting only small errors at the keyframes. The last two rows of Table~\ref{table:keyframing_schemes} show the results for partial keyframes of
root joint trajectory (Root Joint) and VR joints (VR Joints). CondMDI achieves comparable performance on motion
generation while keeping the keyframe error minimal.

Direct comparison of \method{} with SOTA motion diffusion models on the task of keyframe in-betweening is challenging. Some of these models are trained on relative coordinates and thus do not allow for inference-time conditioning on keyframes defined in global coordinates (MDM, PriorMD). GMD is trained with a global coordinate representation, but does not support full keyframes. OmniControl is a recent work that is intended to be used for joint control, but according to the authors, allows for full keyframe conditioning as well. For a more complete comparison, we focus on the task of root joint trajectory control and summarize the performance statistics in Table~\ref{table:root_control}. \method{} demonstrates comparable performance on this task with respect to the SOTA OmniControl model while having a simpler architecture and a relatively faster inference speed. For more details on the inference speed, refer to Appendix~\ref{supp:inference_speed}.

\begin{table}[t]
    \caption{\textbf{Quantitative results for root-joint control} on the HumanML3D test set. \textit{OmniControl (on all)} means the model is trained on all joints.}
    \centering
    \resizebox{0.99\columnwidth}{!}{
    \begin{tabular}{c|ccccc}
        \toprule
        Method & FID $\downarrow$
        & \multicolumn{1}{p{1.9cm}}{\centering R-precision $\uparrow$ \\ (Top-3)}
        & Diversity $\rightarrow$
        &   \multicolumn{1}{p{1.8cm}}{\centering Foot skating \\ ratio $\downarrow$ }
        & \multicolumn{1}{p{1.6cm}}{\centering Keyframe err $\downarrow$}
        \\
        \midrule
        Real & 0.002  & 0.797 & 9.503 & 0.000  & 0.000 \\
        \midrule
        MDM  & 0.698  & 0.602 & 9.197 & 0.1019  & 0.5959 \\
        PriorMDM  & 0.475  & 0.583 & 9.156 & 0.0897  & 0.4417 \\
        GMD       & 0.576  & 0.665 & 9.206 & 0.1009  & 0.1439 \\
        OmniControl (on all)  & \underline{0.322} & \textbf{0.691} & \textbf{9.545} & \textbf{0.0571} & \textbf{0.0367} \\
        \midrule
        Ours  & \textbf{0.2474}  & \underline{0.6752} & \underline{9.4106} & \underline{0.0854} & \underline{0.0525} \\
        \bottomrule
    \end{tabular}
    }
    \label{table:root_control}
\end{table}

\begin{table}[t]
    \caption{\textbf{Ablation results} on the HumanML3D test set. All methods are conditioned on $K=5$ keyframes randomly sampled from the ground truth motion trajectories with the same text prompts in the test set. \textit{IMP} means pure imputation when replacement stops at diffusion step $1$. \textit{C=0} refers to pure imputation with replacement at every diffusion step. \textit{RecG} refers to reconstruction guidance with the default guidance weight ($w_r=20$). \textit{W=5} refers to reconstruction guidance with guidance weight of $w_r=5$. \textit{CondMDI} is our method trained with randomly sampled frames and joints. \textit{(random frames)} denotes training with randomly-sampled full keyframes.}
    \centering
    \resizebox{0.99\columnwidth}{!}{
    \begin{tabular}{c|ccccc}
        \toprule
        Method
        & FID $\downarrow$
        & \multicolumn{1}{p{1.9cm}}{\centering R-precision $\uparrow$ \\ (Top-3)}
        & Diversity $\rightarrow$
        &   \multicolumn{1}{p{1.8cm}}{\centering Foot skating \\ ratio $\downarrow$ }
        & \multicolumn{1}{p{1.6cm}}{\centering Keyframe err $\downarrow$} \\ \midrule
        Real & 0.002  & 0.797 & 9.503 & 0.000  & 0.000 \\
        \midrule
        IMPC=0 & 8.6204 & 0.5710 & 6.3448 & 0.1499  & \textbf{0.0034} \\
        IMP & 0.3600 & \textbf{0.6837} & 9.0170 & 0.1198  &  0.5150 \\
        IMP+RecG  & 1.7072 & 0.6498 & 8.0083 & 0.1720  &  \textbf{0.0034} \\
        IMP+RecGW=5  & 4.4881 & 0.6193 & 7.0836  &  0.1728   & \textbf{0.0034}  \\
        \midrule
        CondMDI (random frames) & {0.1822} & {0.6821} & 9.2648 & 0.0920  &   \underline{0.1165} \\
        CondMDI  &  \textbf{0.1731} &  \underline{0.6823} & \textbf{9.3053} & \textbf{0.0850} & 0.1789\\
        \bottomrule
    \end{tabular}
    }
    \label{table:ablations}
\end{table}

\begin{figure*}
  \resizebox{.8\textwidth}{!}{
     \centering
     \begin{subfigure}[b]{0.24\textwidth}
         \centering
         \includegraphics[width=\textwidth]{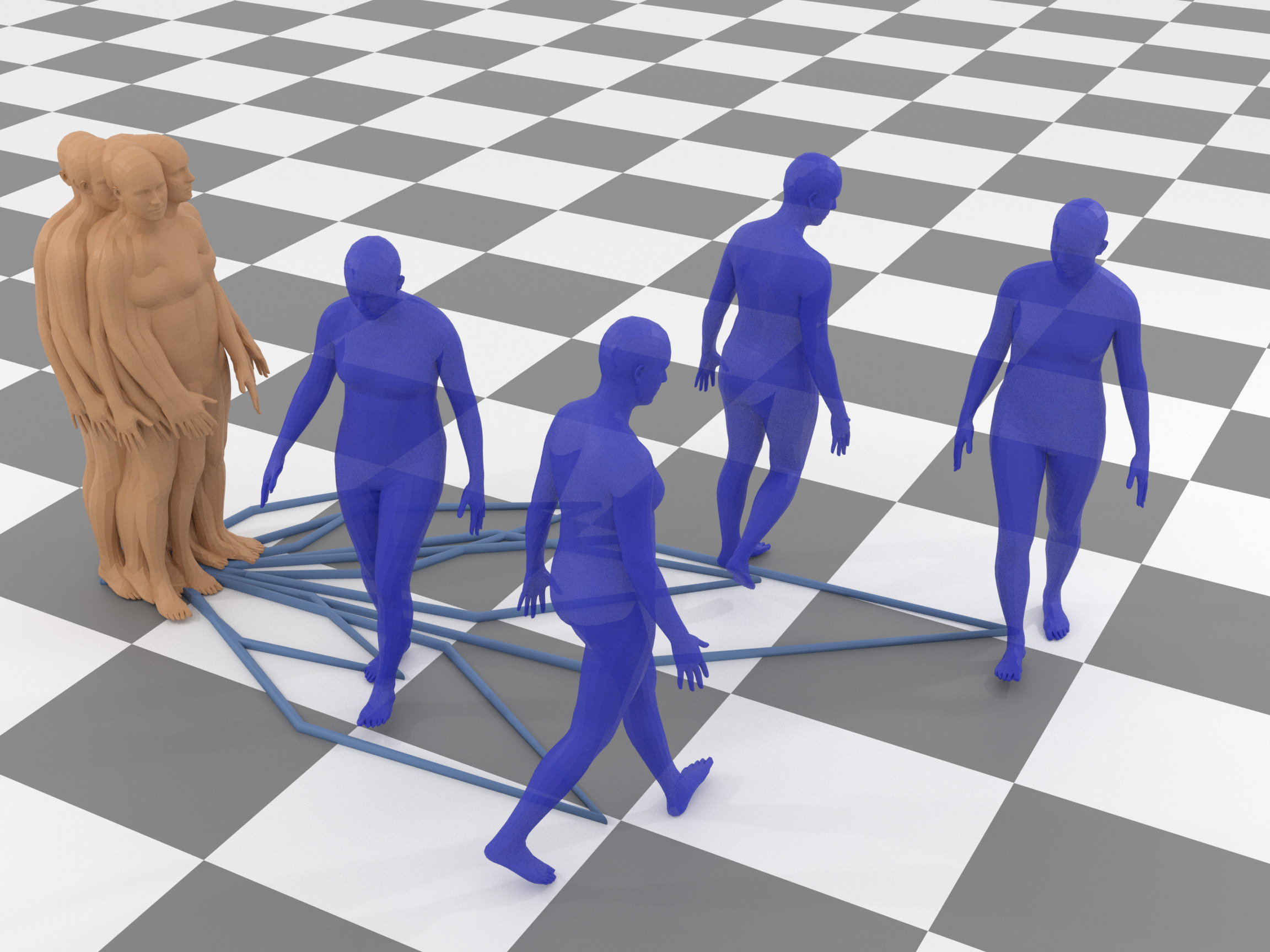}
         \caption{Imputation C=0}
         \label{fig:ablation_imputation_c0}
     \end{subfigure}
     \hfill
     \begin{subfigure}[b]{0.24\textwidth}
         \centering
         \includegraphics[width=\textwidth]{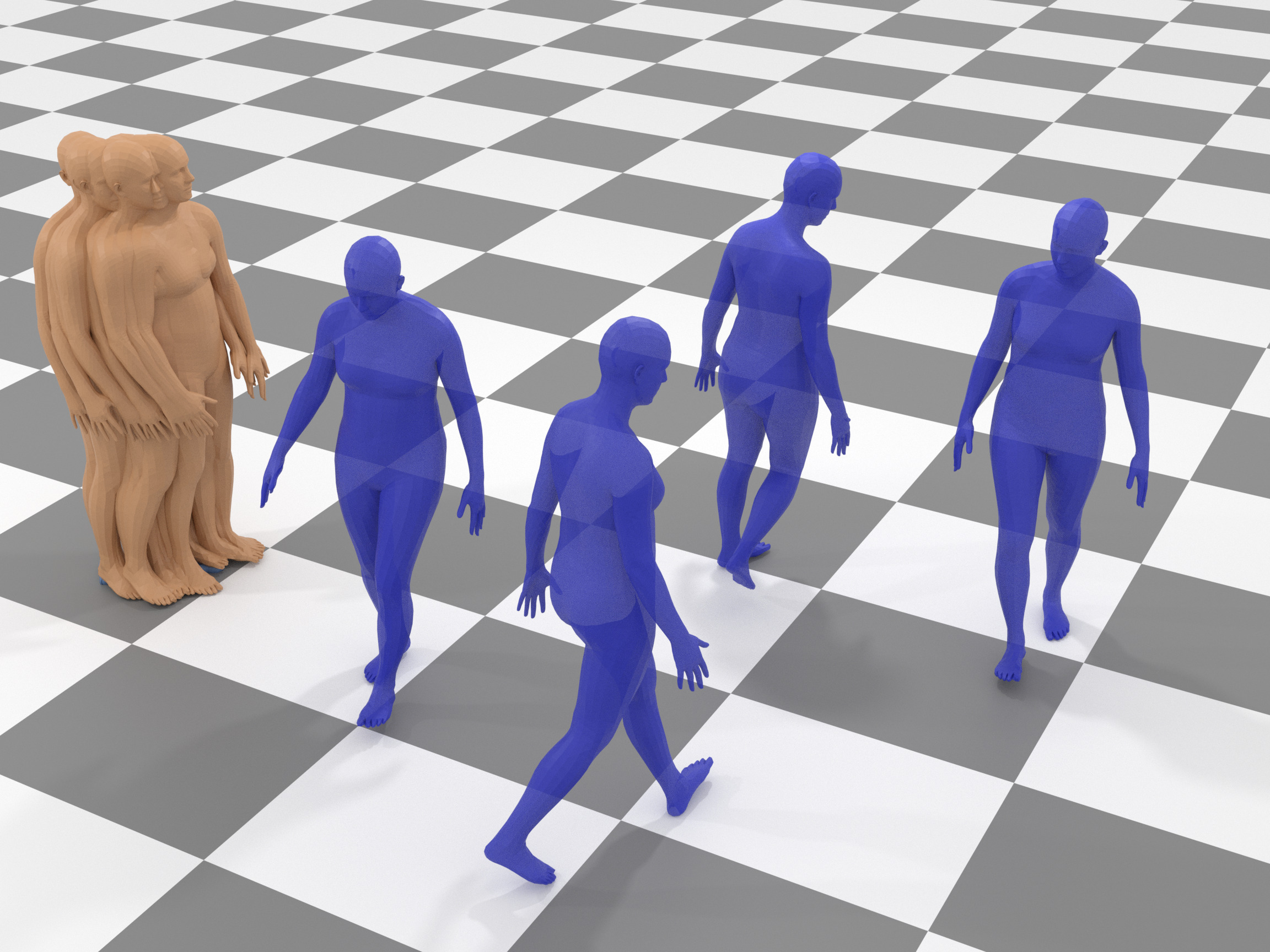}
         \caption{Imputation C=1}
         \label{fig:ablation_imputation_c1}
     \end{subfigure}
     \hfill
     \begin{subfigure}[b]{0.24\textwidth}
         \centering
         \includegraphics[width=\textwidth]{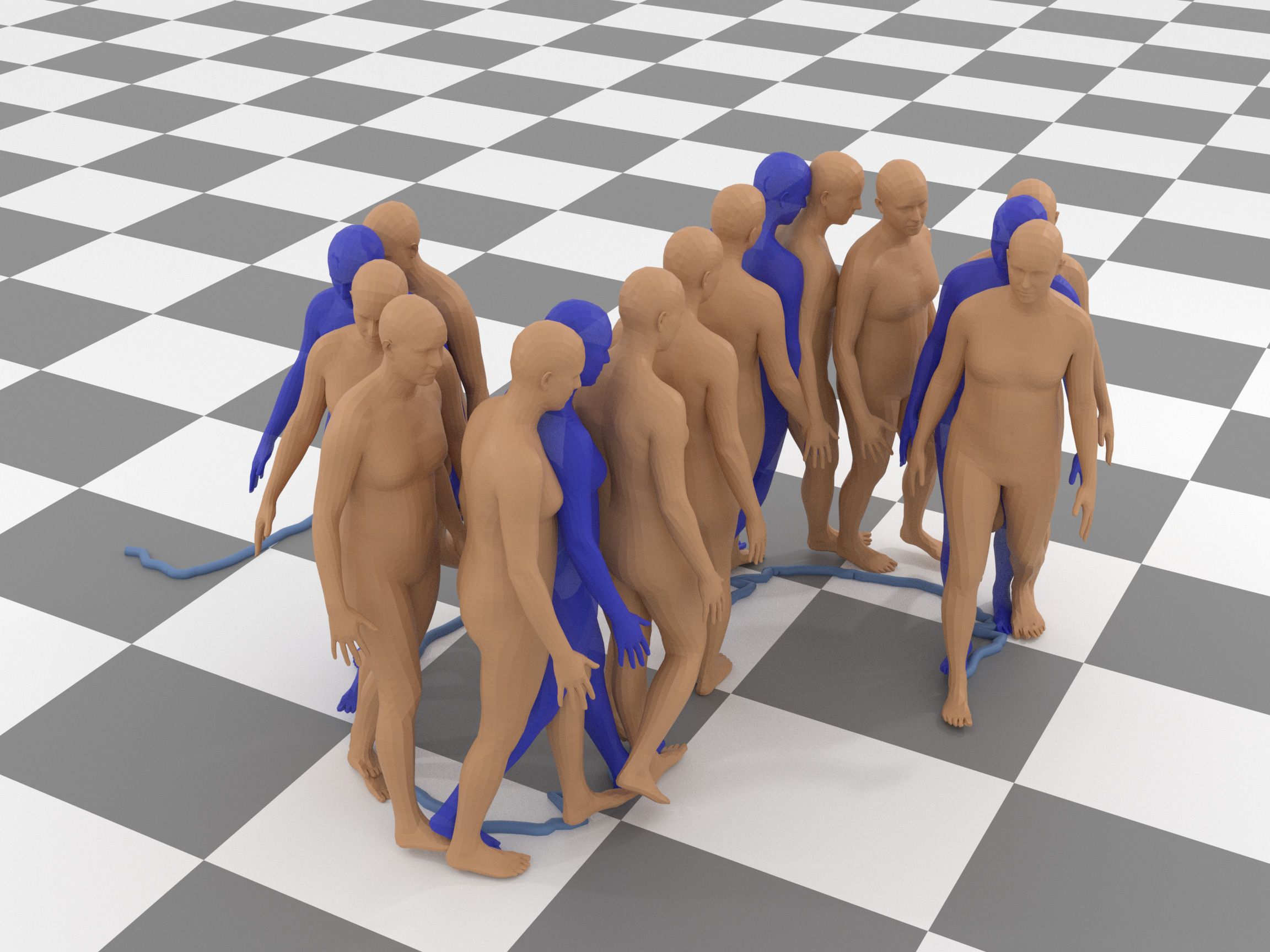}
         \caption{Imputation with guidance}
         \label{fig:ablation_recgui}
     \end{subfigure}
     \hfill
     \begin{subfigure}[b]{0.24\textwidth}
         \centering
         \includegraphics[width=\textwidth]{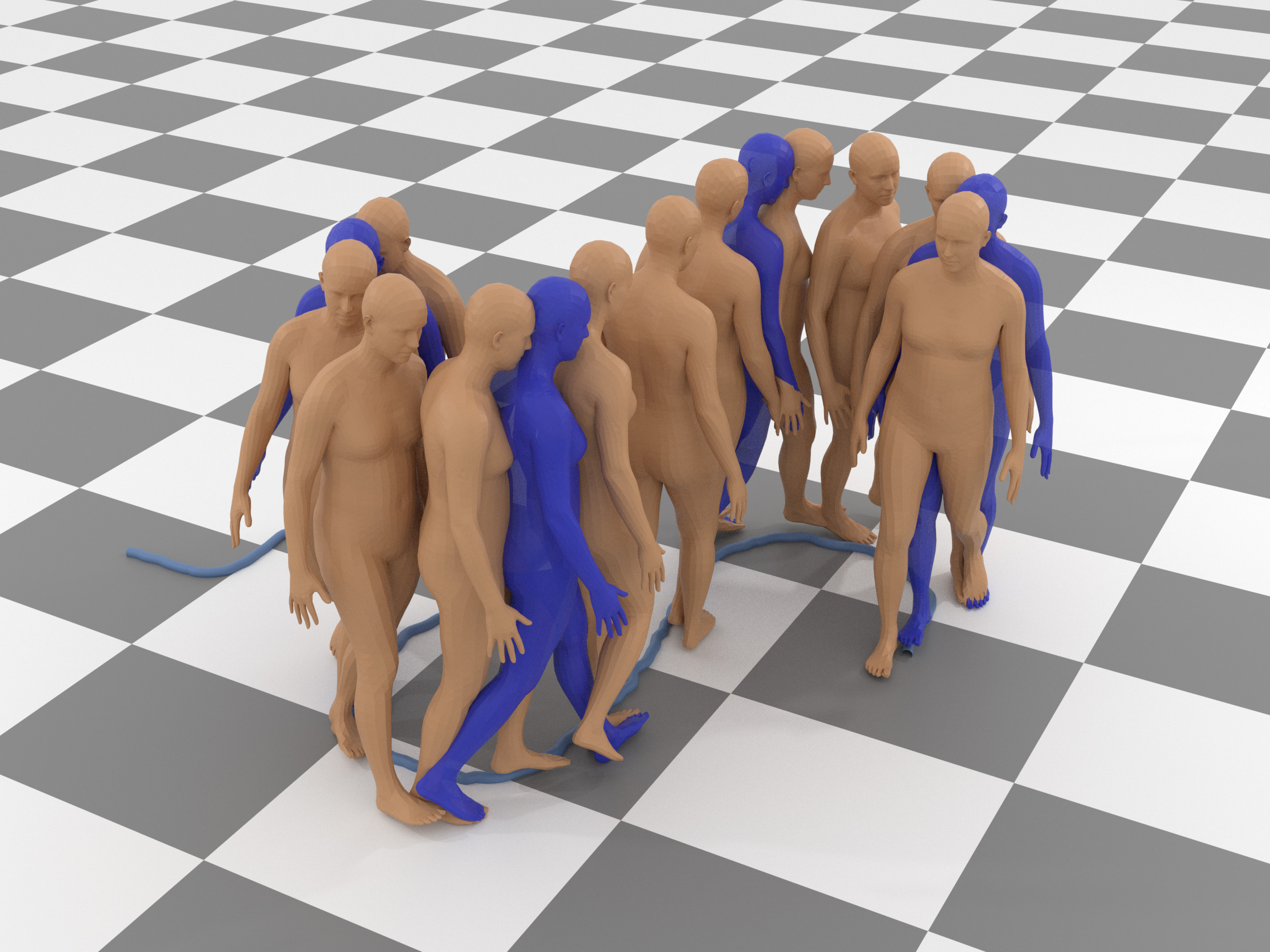}
         \caption{Pure conditional model}
         \label{fig:ablation_conditional}
     \end{subfigure}
    }\\
    \vspace{-10pt}
    \caption{Ablation results on a simple S-walking motion, with keyframes equally spaced $=30$ frames apart. While \emph{Imputation} alone fails to follow the keyframes, \emph{Imputation with guidance} is able to do so but suffers from jitters and inconsistencies. $C$ indicates the denoising step in which replacement stops. For a better look please refer to the supplementary video.}
    \label{fig:ablation_full}
\end{figure*}

\subsection{Ablations}
\label{subsec:ablations}
We perform a comprehensive ablation study over different conditioning methods. Table~\ref{table:ablations} shows the ablations results in which we defined $K=5$ keyframes randomly spaced over motion sequences. Pure imputation which replaces keyframes with ground-truth values at every denoising step (\emph{IMPC=0}) demonstrates minimal error over keyframes, which is expected when replacement is performed until the last denoising step. However, the very large FID score shows that this method leads to unnatural low-quality motions. Figure~\ref{fig:ablation_imputation_c0} shows a sample from this method, which exhibits a large jump before and after every keyframe. This shows that imputation is completely ignored by the diffusion model. Stopping imputation at denoising step 1, (\emph{IMP}) results in high keyframe errors but near-SOTA FID score. Figure~\ref{fig:ablation_imputation_c1} shows such an example for which the model completely ignores the input keyframes but generates a reasonable motion. Adding reconstruction guidance to imputation (\emph{IMP+RecG}) improves both the motion quality metrics and the keyframe-related errors compared to (\emph{IMP}). Figure~\ref{fig:ablation_recgui} shows a sample in which the motion both adheres well to the keyframes while being coherent with the keyframes and the generated frames, reducing the jumps seen with (\emph{IMP}). Finally, \method{} exhibits the best performance compared to inference-conditioning methods. Figure~\ref{fig:ablation_conditional} shows a smooth motion that adheres closely to the keyframes.

Finally, we perform an ablation study over the choices of random mask generation
schemes used during training. In Table~\ref{table:ablations}, \emph{(random frames)} correspond to our model trained with keyframes generated by randomly sampling the number and location of observed keyframes, while always including all the joints. Although this model has comparable FID scores and improved keyframe error compared to \method{} on this task, it does not generalize well to partial keyframes. In general, \method{} does better on partial keyframe in-betweening tasks, as the model is trained with partial keyframes.
\section{Conclusion}
\label{sec:conclusion}

We have presented a simple and flexible diffusion-based method for keyframe motion completion.
It allows for flexibility at inference time and has motion quality comparable to the current state-of-the-art for diffusion-based models. Our method can be used with any backbone motion diffusion model with minimal changes and can therefore readily take advantage of continuing improvements there.
We demonstrate our mask-conditioned method with sparse and dense keyframes, partial keyframes, and 
text conditioning, and show its ability to generate diverse samples.
In addition to comparing to related work on the HumanML3D dataset, we give empirical results for several ablations and alternative inference-time conditioning variations. 

Our work comes with a number of limitations and related future work.
The resulting motions still exhibit some minor footskate and motion jitter for highly dynamic motions, which could likely
be addressed with an appropriate footskate or smoothness loss or by leveraging a physics-based simulation
to track the generated motion. 
The HumanML3D dataset used for training includes skating and swimming data, 
and thus removing these outlier motions from the dataset, or providing extra contextual information
about them, may also help reduce remaining footskate artefacts. 
Our current keyframe selection algorithm used during training is fully randomized. We are interested in improving the algorithm by grounding it to combinations that are most used in practice.
Finally, our model works with keyframes represented with the same representation as the HumanML3D dataset. This redundant data representation introduces a challenge when conditioning on partial keyframes because spatial constraints may correspond to a small number of features, resulting in the model treating these sparse observed values as noise. Therefore, we are interested in extending our framework to address the issues resulting from uneven representation of different features. 
\begin{acks}
We thank Saeid Naderiparizi for his valuable insights and feedback during the early stages of the work. This work was supported by the NSERC grant RGPIN-2020-05929 and was enabled in part by technical support and computational resources provided by Digital Research Alliance of Canada (\url{www.alliancecan.ca}).
\end{acks}

\bibliographystyle{ACM-Reference-Format}
\bibliography{main}

\clearpage
\appendix

\twocolumn[{%
 \centering
 \Huge Appendix\\[1.5em]
}]

\section{Motion Representation Details}
\label{supp:represenation}
Our method assumes the motion $\rvx \in \mathbb{R} ^ {N \times J \times D}$ to include a sequence of poses over $N$ frames, where the pose in each frame consists of $J$ joints, where each represented by $D$ features. In the HumanML3D dataset~\cite{guo2022generating}, motion sequences have variable lengths between 39 and 196 frames. Thus, shorter motions are padded with zeros such that $N=196$ for all motions. Each frame is represented with a $263$-dimensional feature vector thus $J=263$ and $D=1$ for this particular representation.

The human motion representation used in HumanML3D dataset follows the convention of dividing the the motion into two parts: \emph{local motion}, which contains the pose of the skeleton relative to the root at every frame, and \emph{global motion}, which contains the global translation and rotation of the root joint relative the the previous frame. Therefore, the representation of the motion at frame $t$ can be shown as below:
\begin{align}
    \rvx_t = \left< \rvx^{\texttt{global}}_t, \rvx^{\texttt{local}}_t\right> \in \mathbb{R}^{263}
\end{align}
where $\rvx^{\texttt{local}}_t$ and $\rvx^{\texttt{global}}_t$ represent the \emph{local} and \emph{global} motions at frame $t$ respectively. The \emph{global} motion at time $t$ is composed of the relative root rotation with respect to the previous frame $\dot{{\theta}}_t$, relative x and z displacement of the root joint with respect to the previous frame $\dot{\rvr}_t$, and the root joint height ${r}^{\texttt{h}}_t$:
\begin{align}
    \rvx^{\texttt{global}}_t = \left< \dot{\mathbf{\theta}}_t, \dot{\rvr}_t, {r}^{\texttt{h}}_t \right> \in \mathbb{R}^{4}.
\end{align}
The \emph{local} motion at time $t$ is composed of the local joint positions with respect to the root $\rvx^p_t \in \mathbb{R}^{21 \times 3}$, the local joint rotations with respect to the root $\rvx^r_t \in \mathbb{R}^{21 \times 6}$, the global joint velocities $\dot{\rvx}^p_t \in \mathbb{R}^{22 \times 3}$ and the foot contact information $\rvc_t \in \mathbb{R}^{4}$:
\begin{align}
    \rvx^{\texttt{local}}_t = \left< \rvx^p_t , \rvx^r_t, \dot{\rvx}^p_t, \rvc_t \right> \in \mathbb{R}^{259}
\end{align}
where the number of joints in the dataset is 22.

\section{Global vs. Relative Root Representation}
\label{supp:root_represenation}
To make the keyframe definition more intuitive and keyframe-conditioning more straight-forward, we change the root data representation from relative (with respect to the previous frame) rotation and position to global (absolute). To covert the \emph{global} part of the motion to global, for every frame, we simply sum the rotations and positions of the root joint in all frames before it. Therefore, the final dataset that our model is trained on changes the \emph{global} motion as below while keeping the rest of the features the same:
\begin{align}
    \rvx^{\texttt{global}}_t = \left< {\mathbf{\theta}}_t, {\rvr}_t, {r}^{\texttt{h}}_t \right> \in \mathbb{R}^{4}.
\end{align}
It is worth noting that for GMD, it has been demonstrated that this change in root representation does not negatively impact the performance of motion generation \cite{karunratanakul2023gmd}. Therefore, we make this adjustment confidently.

\section{Keyframe Signal Details}
\label{supp:keyframe_signal_details}
Our method assumes that the keyframe signal and motion signal have the same dimensionality, thus for every observed frame and observed joint, \method{} requires all the corresponding features out of 263. For instance, when conditioned on the root joint trajectory, our method observes the first 4 values of the feature vector for every frames. This also means that conditioning on partial keyframes requires the observation of the root joint, as the other joints are represented with respect to the root joint in this particular dataset. Foot contact information will be available to the model only if the corresponding foot or ankle joints are observed.

\section{Implementation Details}
\label{supp:implementation_details}
We implemented our model using PyTorch and used the unconditioned motion diffusion model of GMD~\cite{karunratanakul2023gmd} as the backbone of \method{}.

\subsection{Network Architecture}
\label{supp:architecture}
Following~\citet{karunratanakul2023gmd}, our motion diffusion model is a UNet with 1D convolutions and Adaptive Group Normalization (AdaGN)~\cite{dhariwal2021diffusion}. We present the choice of hyperparameters in Table~\ref{table:hyperparameters}.

\begin{table}[h]
    \caption{\textbf{Hyperparameters.}}
    \centering
    \resizebox{0.75\columnwidth}{!}{
    \begin{tabular}{c|c}
        \toprule
        Hyperparameter & Value
        \\
        \midrule
        Training iterations & 1M \\
        Learning rate & 1e-4 \\
        Optimizer & Adam W \\
        Weight decay & 1e-2 \\
        Batch size & 64 \\
        Channels dim & 512 \\
        Channel multipliers & $[2,2,2,2]$ \\
        Variance scheduler & Cosine~\cite{nichol2021improved} \\
        Diffusion steps & 1000 \\
        Diffusion variance & $\tilde{\beta} = \frac{1-\alpha_{t-1}}{1-\alpha_t}\beta_t$ \\
        EMA weight ($\beta$) & 0.9999 \\
        \bottomrule
    \end{tabular}
    }
    \label{table:hyperparameters}
\end{table}

\subsection{Training Details}
\label{supp:training}
We used a batch size of 64 and trained our model on a single NVIDIA A100 GPU. Our inference-time in-betweening model (GMD's unconditioned motion diffusion model) is trained for 500K iterations, while \method{} is trained for 1M iterations. We use Adam W optimizer~\cite{loshchilov2017decoupled} with a learning rate of $0.0001$ and weigh decay of $0.01$. Following \citet{karunratanakul2023gmd}, we do not use dropout, and we clip the gradient norm to 1 for increased training stability. We used the exponential moving average (EMA) of trained snapshots of the model during training ($\beta = 0.9999$), and used the average model for better generation quality.

\subsection{Inference Details}
\label{supp:inference}
We use a value of $w=2.5$ for the classifer-free guidance weight. The the inference-time conditioning methods are implemented by slightly modifying the sampling algorithm of \method{}. For imputation, we replace the observed parts of the sample estimate $hat{\rvx}_0$ with the keyframes over the observation mask $m$. For reconstruction guidance, we guide the unobserved parts of the sample estimate $\hat{\rvx}_0$ using the gradient of the keyframe reconstruction loss. Algorithm~\ref{alg:full_sampling} shows an overview of the sampling procedure used for the inference-time methods in the ablations.

\SetKwComment{Comment}{\quad $\triangleright$ }{ }
\begin{algorithm}[h]
\caption{Sampling: Inference-time In-betweening} \label{alg:full_sampling}
\small
\textbf{Require:} Guidance scale $w$  \\
\textbf{Require:} Text prompt $\rvp$  \\
\textbf{Require:} Keyframe signal $\rvc$ and observation mask $\rvm$  \\
$\rvx_T \sim \mathcal{N}(\vzero, \mI)$ \\
\For{$t=T, \dotsc, 1$}{
    $\rvz \sim \mathcal{N}(\vzero, \mI)$ if $t > 1$, else $\rvz = \vzero$ \\
    $\hat{\rvx}_0 = G_\theta(\rvx_t,t,\emptyset) + w(G_\theta(\rvx_t, t, \rvp) - G_\theta(\rvx_t,t,\emptyset))$ \\
    \If{reconstruction guidance}{
        {$\tilde{\rvx}_0 = \hat{\rvx}_0 - \frac{w_r \sqrt{\bar{\alpha}_t}}{2}\nabla_{\rvx_t} \left\| {\rvc - \hat{\rvx}_0}\right\|^2$} \\
        {$\hat{\rvx}_0 = \rvm \odot \hat{\rvx}_0 + (\mathbf{1} - \rvm) \tilde{\rvx}_0$}
    }
    \If{imputate}{
        {$\hat{\rvx}_0 = \rvm \odot \rvc + (\mathbf{1} - \rvm) \odot \hat{\rvx}_0$}
    }
    $\hat{\vmu} = \tilde{\vmu}(\hat{\rvx}_0, \rvx_t)$ \\
    $\rvx_{t-1} = \hat{\vmu} + \sigma_t \rvz$
}
\textbf{Return} $\rvx_0$
\end{algorithm}

\section{Comparison with GMD}
 For the motion diffusion model of \method{}, we use GMD's second-stage model without their trajectory conditioning and emphasis projection. We chose this architecture as we required training \method{} on motion data with global-root representations as described in Section~\ref{subsec:motion_representation}. MDM~\cite{tevet2023human}, a Transformer-based text-conditioned motion diffusion model, shows a significant drop in performance when trained with the global-root representation data. In contrast, GMD’s motion diffusion model, a UNet-based text-conditioned motion diffusion model, has comparable performance for the new representation and the original relative representation~\cite{karunratanakul2023gmd}. The used backbone diffusion model is only capable of text-conditioned motion generation and does not support any spatial conditioning without our approach added to it.

\section{Text-to-Motion Evaluation Metrics}
\label{supp:metrics}
Originally suggested by~\citet{guo2022generating}, the following metrics are based on a text feature extractor and a motion feature extractor jointly trained under a contrastive loss to produce geometrically close feature vectors for matched text-motion pairs, and vise versa.

\textbf{R Precision (top-3).}
Evaluates the relevance of a generate motion and its text prompt. To compute this metric, we first create a batch of generated motions with their ground-truth text from the test set. For each generated motion in the batch, we calculate the euclidean distance between the motion feature and every text feature within the batch. We then sort the texts based on their distances to each motion. If the ground-truth text falls into the top-3 candidates, we treat this motion as a true positive retrieval and give this motion a score of 1., and a 0. otherwise. Final metric is computed as the average score within all motions of all batches. We use batch size $32$ (i.e. 31 negative text descriptions per motion).

\textbf{Fréchet Inception Distance (FID).} Is a widely used metric to evaluate the overall quality for generation tasks. This metric is computated as the distance between a large set of generated samples vs. ground-truth samples from the test set. To compute the distance, all generated and ground-truth samples are first fed into a feature extractor network (the Inception network \cite{szegedy2015going} for image generation) from which the features are extracted. Then, a Gaussian distribution is fitted to each set of features. The FID score is computed as the Fréchet distance between the two Gaussian distributions which can be solved in closed-form. To compute this metric, we generate 1000 motions, and use the evaluator network provided by \cite{tevet2023human} as the feature extractor.

\textbf{Diversity.} Measures the variance of the generated motions across all action categories. To compute this metric, we first randomly sample two subsets of the same size $S_d$ out of the set of all generated motions across all action categories denoted $\{\rvv_1,...,\rvv_{S_{d}}\}$ and $\{\rvv'_1,...,\rvv'_{S_{d}}\}$.
The diversity of those sets of motions is defied as
\begin{equation}
    \mathrm{Diversity} = \frac{1}{S_d}\sum_{i=1}^{S_d}\parallel \rvv_i-\rvv_i' \parallel_2.
\end{equation}
We use $S_d = 200$ for our experiments. The diversity value is considered better when closer to the diversity value of the ground truth.

\section{Inference Speed}
\label{supp:inference_speed}
We report the inference speed of our method, and baseline methods in Table~\ref{table:inference_speed}. The infrence time is computed by defining sparse keyframes every $T=20$ frames for a single motion examples averaged over $10$ trials. Experiments where all performed on an NVIDIA GeForce RTX 2070 GPU.

\begin{table}[h]
    \caption{\textbf{Inference time.}}
    \centering
    \resizebox{0.8\columnwidth}{!}{
    \begin{tabular}{c|cccc}
        \toprule
        Method & Ours & MDM & GMD & OmniControl
        \\
        \midrule
        Time (s) & 54.39 $\pm$ 0.59  & 59.30 $\pm$ 0.53 & 166.42 $\pm$ 0.98 & 183.79 $\pm$ 0.73 \\
        \bottomrule
    \end{tabular}
    }
    \label{table:inference_speed}
\end{table}

\section{Additional Results}
\label{supp:additional_results}
\subsection{Diversity}
\begin{figure*}[t]
  \resizebox{.9\linewidth}{!}{
     \centering
     \begin{subfigure}[b]{0.16\textwidth}
         \centering
         \includegraphics[width=\textwidth,trim={300 150 120 0},clip]{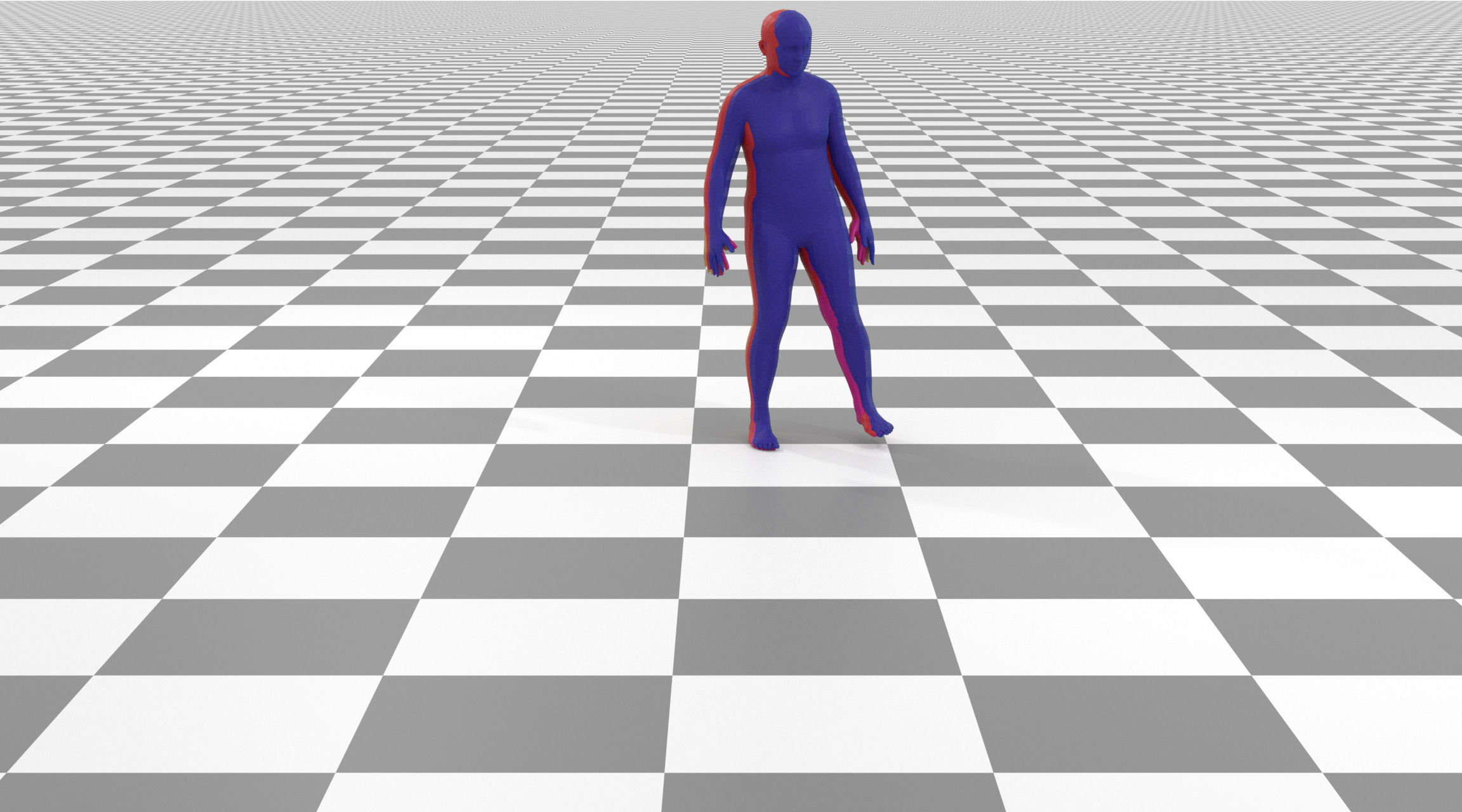}
         \caption{}
     \end{subfigure}
     \hfill
     \begin{subfigure}[b]{0.16\textwidth}
         \centering
         \includegraphics[width=\textwidth,trim={300 150 120 0},clip]{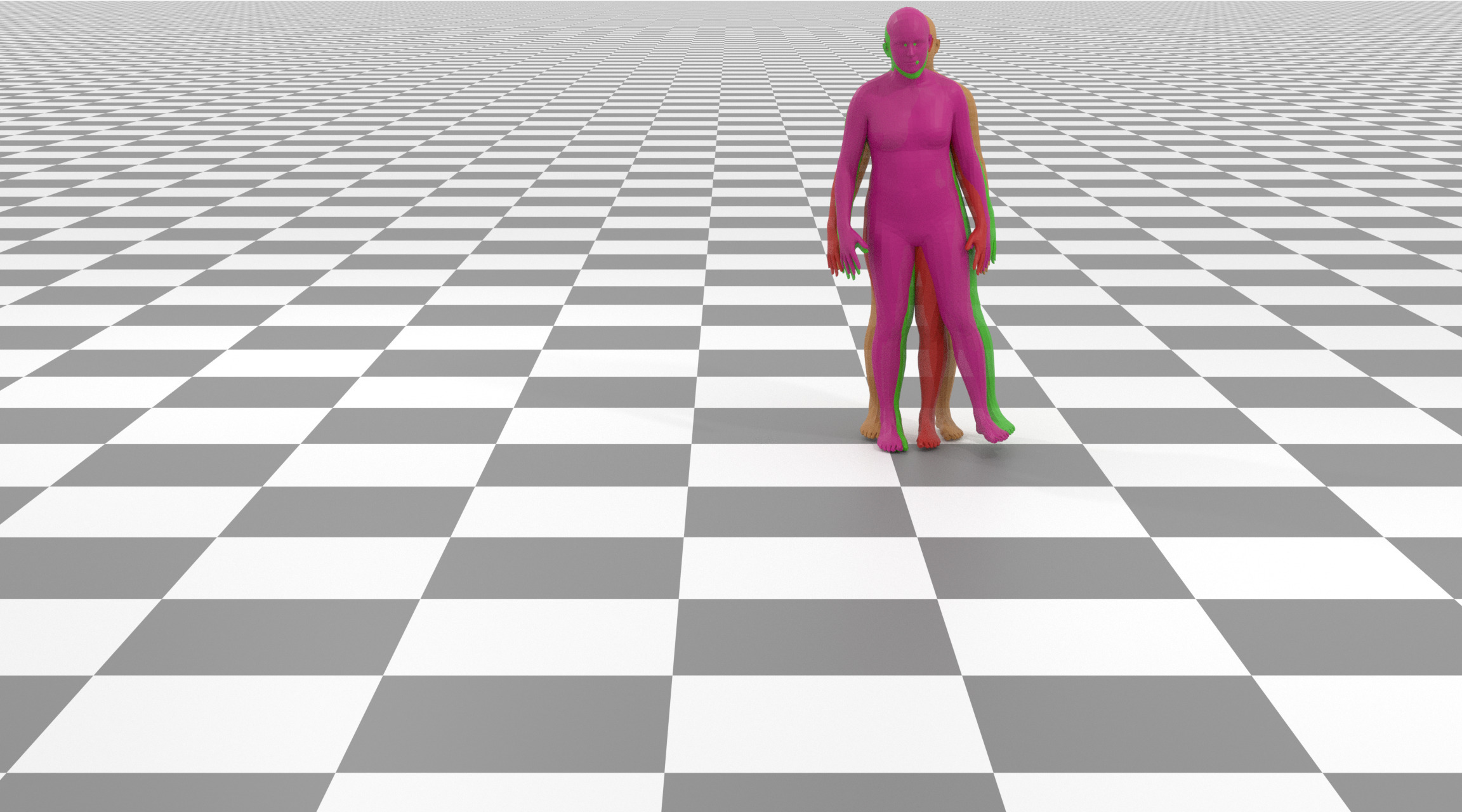}
         \caption{}
     \end{subfigure}
     \hfill
     \begin{subfigure}[b]{0.16\textwidth}
         \centering
         \includegraphics[width=\textwidth,trim={300 150 120 0},clip]{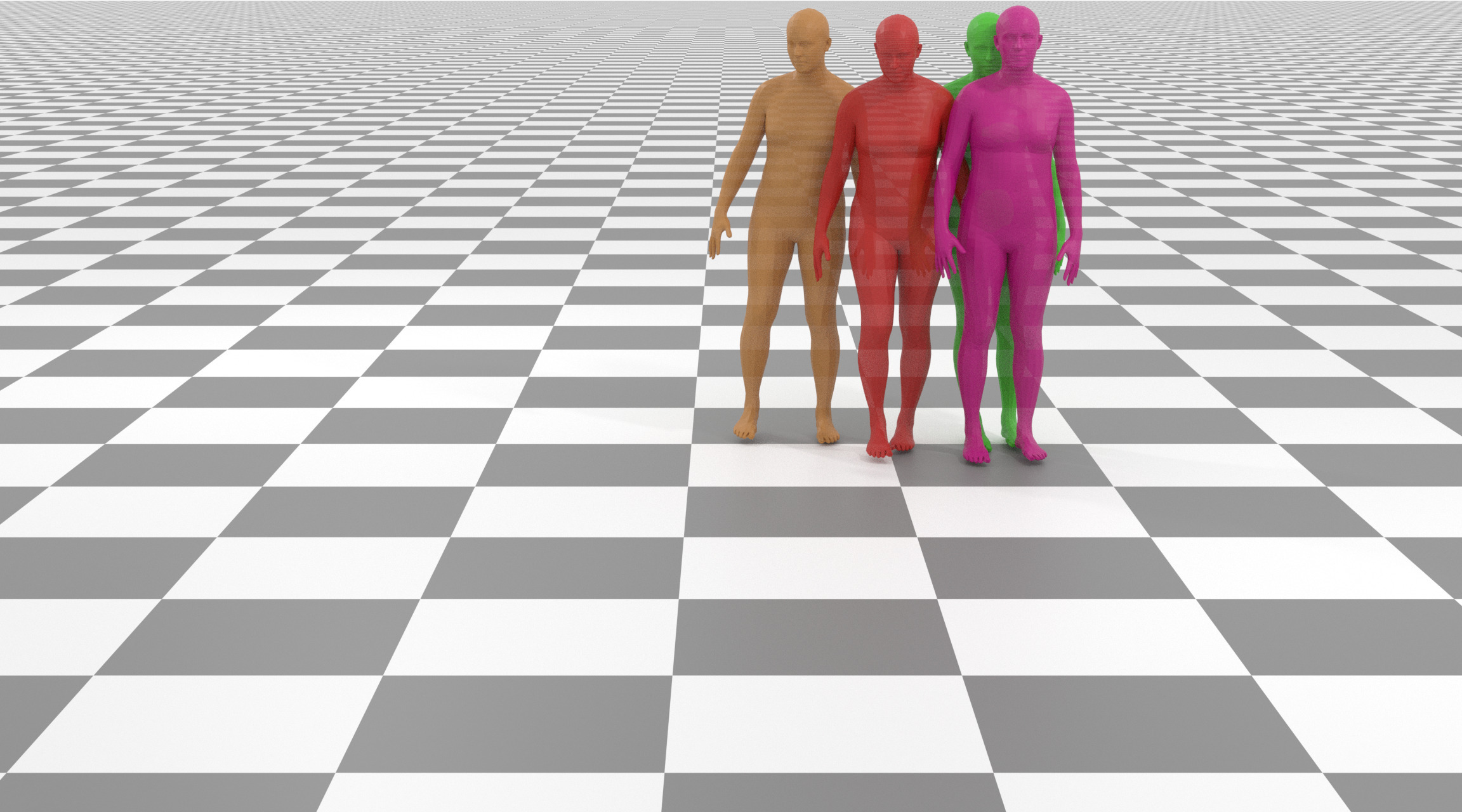}
         \caption{}
     \end{subfigure}
     \hfill
     \begin{subfigure}[b]{0.16\textwidth}
         \centering
         \includegraphics[width=\textwidth,trim={300 150 120 0},clip]{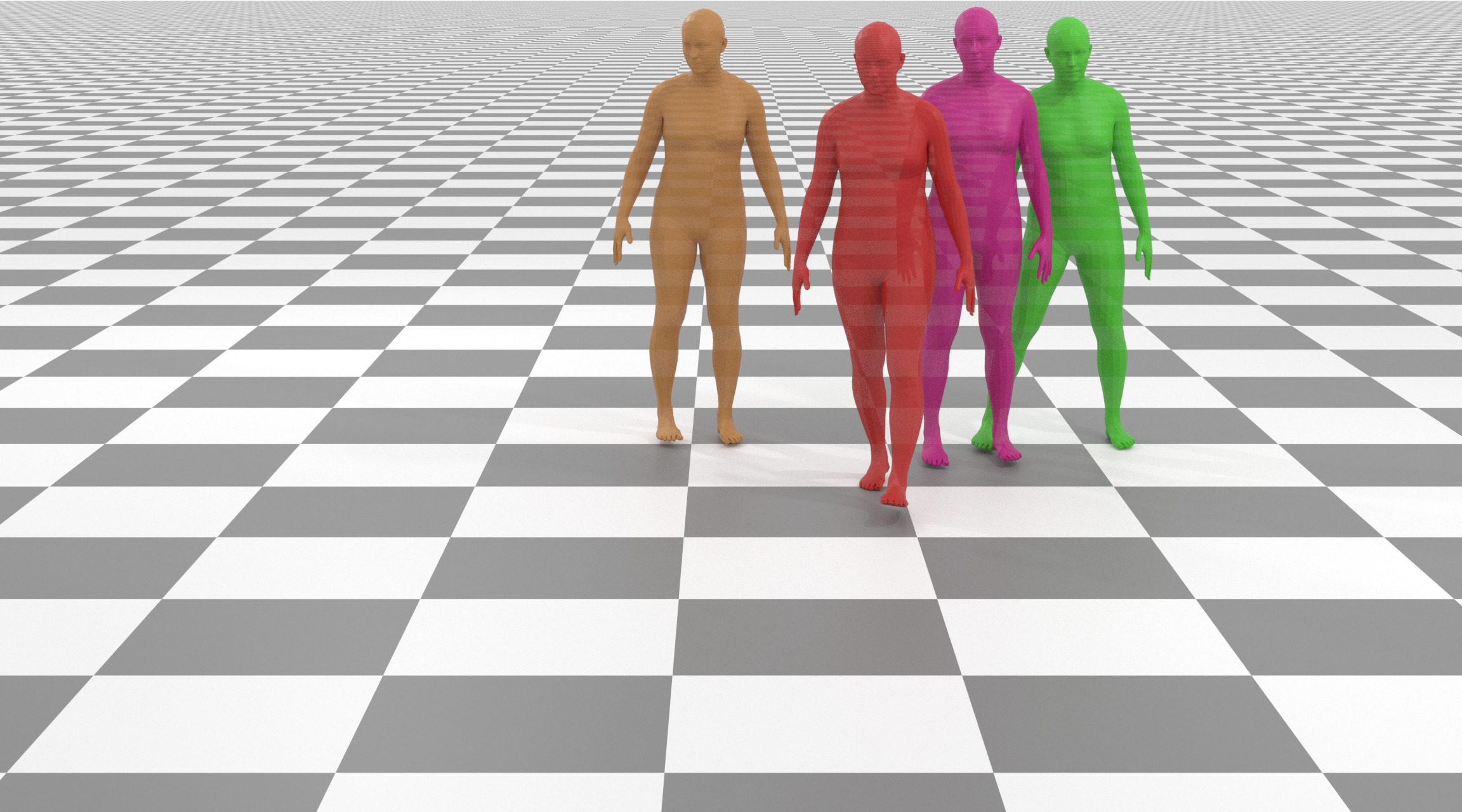}
        \caption{}
     \end{subfigure}
     \hfill
     \begin{subfigure}[b]{0.16\textwidth}
         \centering
         \includegraphics[width=\textwidth,trim={300 150 120 0},clip]{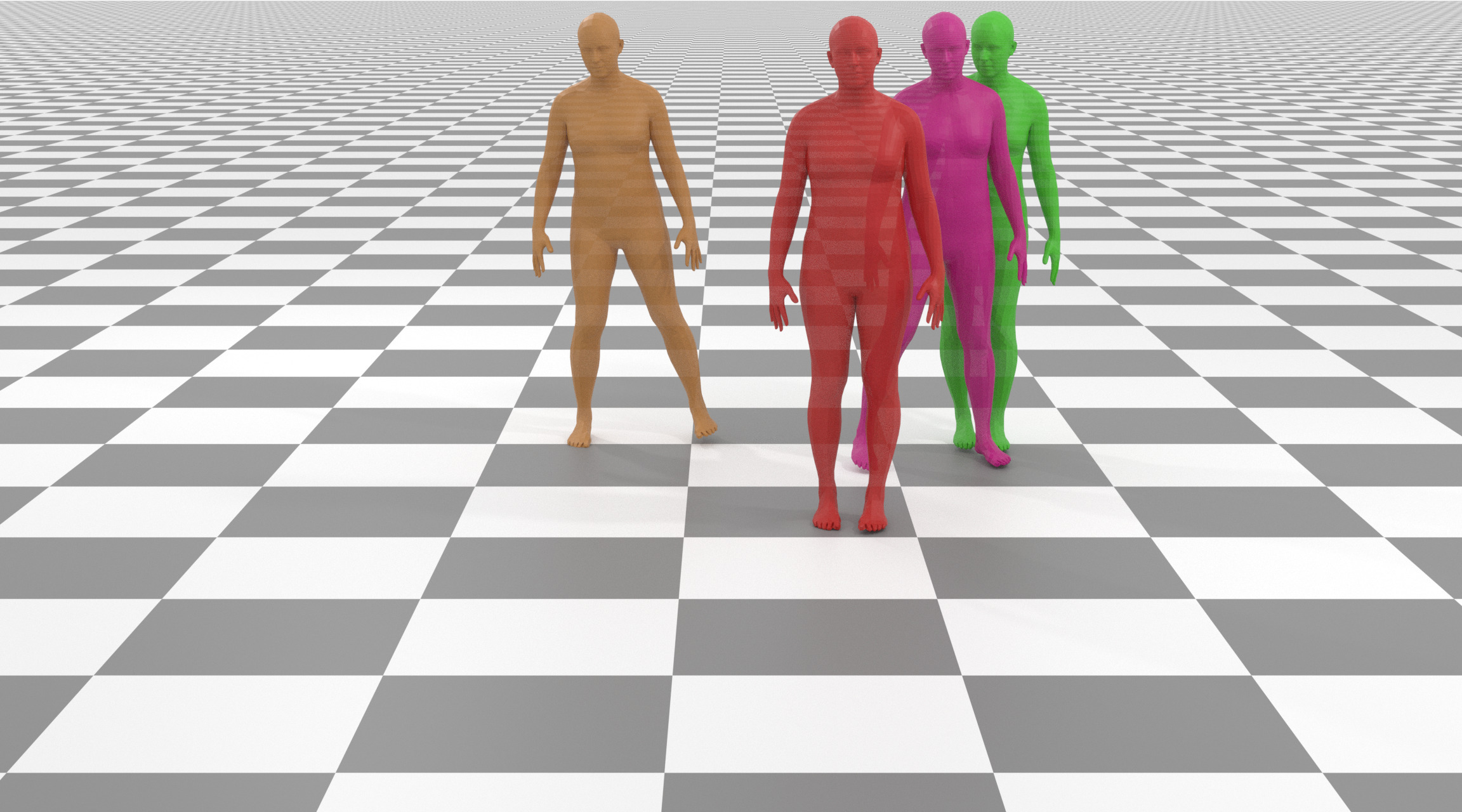}
         \caption{}
     \end{subfigure}
     \hfill
     \begin{subfigure}[b]{0.16\textwidth}
         \centering
         \includegraphics[width=\textwidth,trim={300 150 120 0},clip]{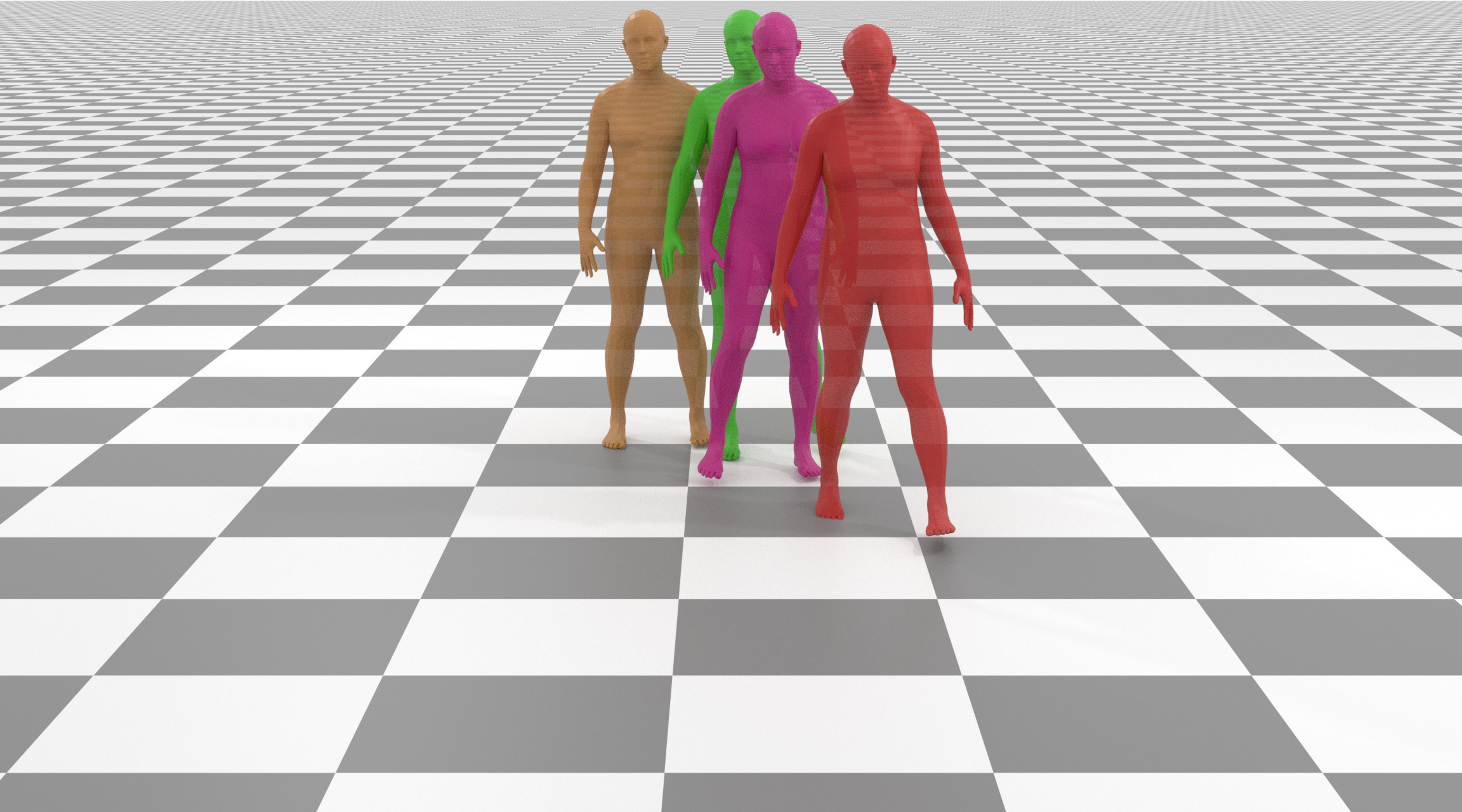}
         \caption{}
     \end{subfigure}
     }
    \vspace{-10pt}
    \caption{Different motions generated with the same conditioning keyframes. After the last keyframe in blue in (a), the motions (displayed in different colors) show diverse and coherent behavior over time (from left to right). Please refer to the supplementary video for a dynamic version with more samples.}
    \label{fig:diversity_strip}
\end{figure*}

Our model displays diverse outputs while also staying cohesive to the keyframes. To show diverse outputs, we condition the model only on 4 keyframes at the beginning of the motion, at timesteps 0, 10, 20, 30, and the model is then free to generate the subsequent motion unconstrained. In Figure~\ref{fig:diversity_strip}, we present the last keyframe, and 4 different samples (in different colors) with different behaviours over future frames.

\subsection{Text conditioning}
\begin{figure}[h]
  \resizebox{\linewidth}{!}{
    \centering
    \includegraphics[width=\linewidth,trim=10 150 10 80,clip]{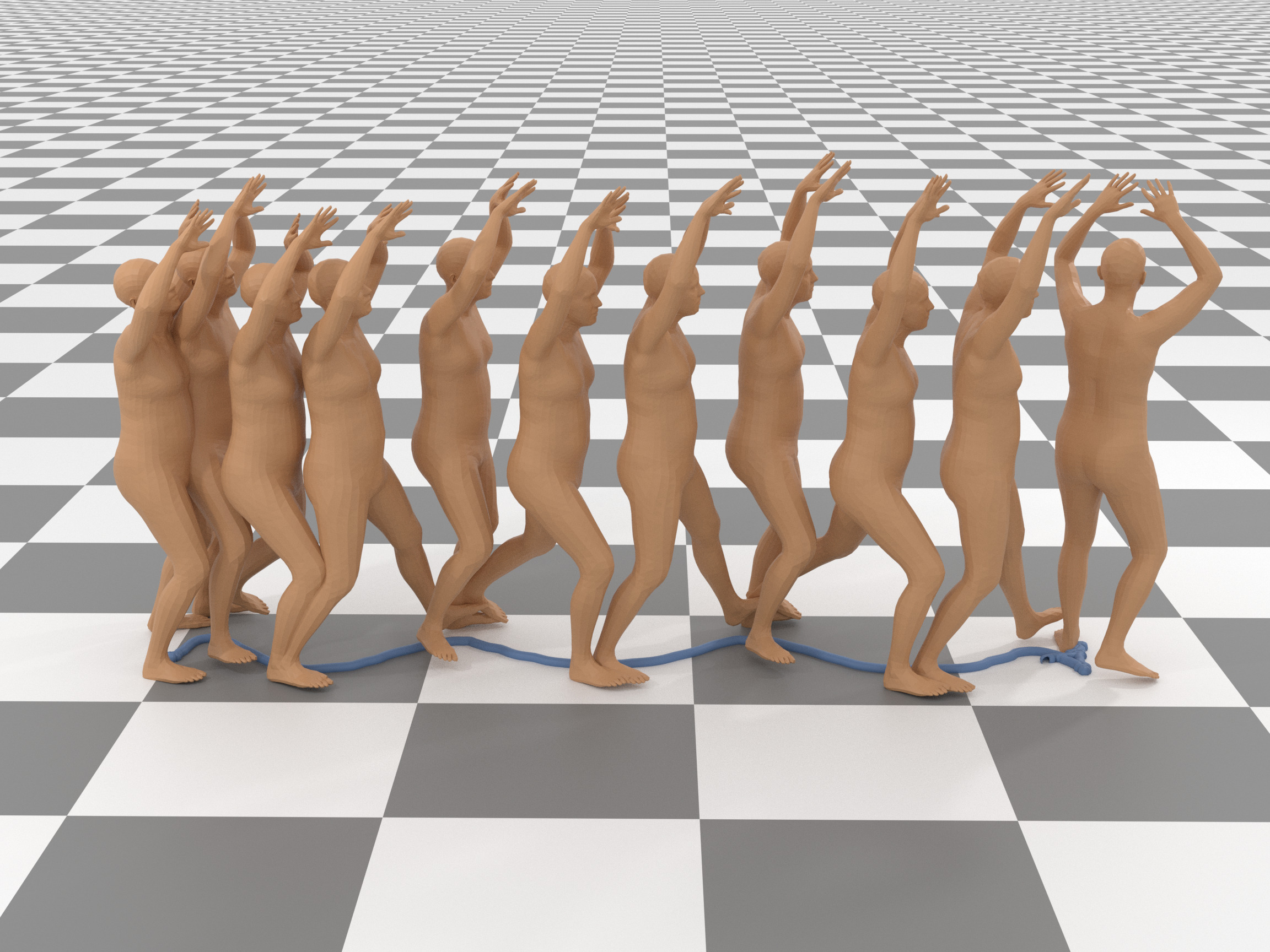}
    \includegraphics[width=\linewidth,trim=60 120 60 25,clip]{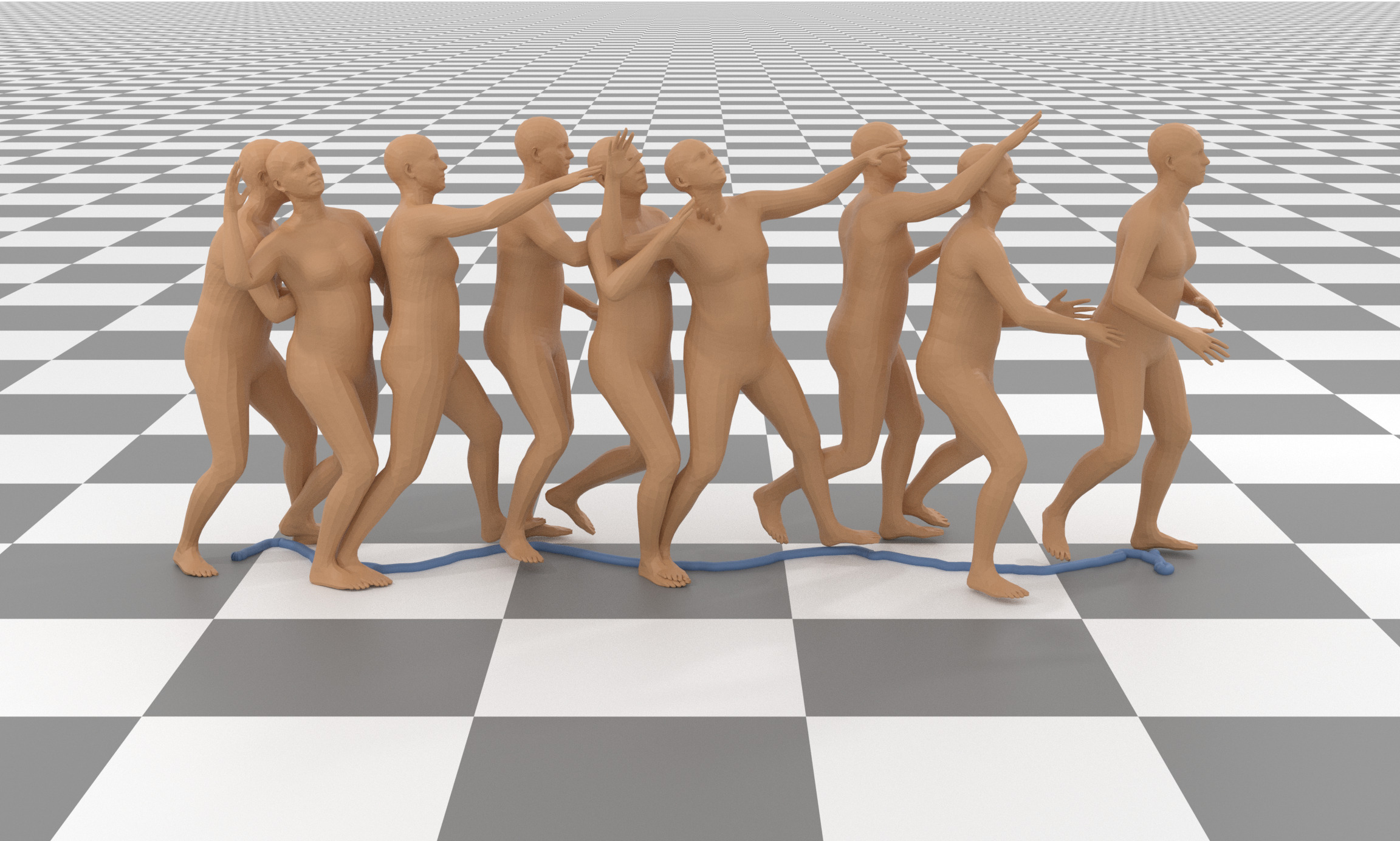}
    }
    \caption{A walking motion conditioned only on the root joint trajectory (projected on the ground) and guided with text "a person is waving their hands above their head" on the left and "a person tosses a ball" on the right.}
    \label{fig:text-conditioning}
\end{figure}

Text conditioning allows the user to guide the output towards specific motions at inference time. It is especially useful when the model is conditioned only on a subset of joints, allowing for more flexibility on the generated motions. Figure~\ref{fig:text-conditioning} shows two examples where the model is conditioned only on the root trajectory provided with two different text prompts, leading to distinct behaviors that follow the same underlying trajectory.

\end{document}